\documentclass[11pt]{article}
\date{}

\usepackage[utf8]{inputenc} 
\usepackage[T1]{fontenc}    
\usepackage[colorlinks=true,linkcolor=blue,allcolors=blue]{hyperref}      
\usepackage{url}           
\usepackage{booktabs}     
\usepackage{amsfonts}   
\usepackage{nicefrac}       
\usepackage{microtype,nicefrac}     
\usepackage{xspace}
\usepackage[numbers]{natbib}
\usepackage{wrapfig}
\usepackage{comment}
\usepackage{amsmath}
\usepackage{amsthm}
\usepackage{graphicx}
\usepackage{rotating}
\usepackage{caption}
\usepackage{subcaption}
\usepackage{amssymb}
\usepackage{autobreak}
\usepackage{mathtools}
\usepackage[dvipsnames]{xcolor}
\usepackage{bbm}
\usepackage[ruled,vlined, linesnumbered]{algorithm2e}
\usepackage{algorithmic}
\usepackage[parfill]{parskip}
\usepackage{authblk}
\usepackage{nicefrac}      
\usepackage{microtype}     
\usepackage[dvipsnames]{xcolor}         
\usepackage{float}
\usepackage{amssymb}
\usepackage{mathtools}

\usepackage{scrwfile}

\TOCclone[\contentsname~(\appendixname)]{toc}{atoc}
\newcommand\StartAppendixEntries{}
\AfterTOCHead[toc]{%
  \renewcommand\StartAppendixEntries{\value{tocdepth}=-10000\relax}%
}
\AfterTOCHead[atoc]{%
  \edef\maintocdepth{\the\value{tocdepth}}%
  \value{tocdepth}=-10000\relax%
  \renewcommand\StartAppendixEntries{\value{tocdepth}=\maintocdepth\relax}%
}
\newcommand*\appendixwithtoc{%
  \cleardoublepage
  \appendix
  \addtocontents{toc}{\protect\StartAppendixEntries}
  \listofatoc
}

\makeatletter

\newcommand{\weight}{w}

\newcommand{\FuncClass}{\mathbb{F}}

\newcommand{\G}{\mathcal{G}}

\newcommand{\CDF}{\textsc{{CDF}}\xspace}

\newcommand{\WIS}{\textsc{WIS}\xspace}
\newcommand{\IS}{\textsc{IS}\xspace}
\newcommand{\ISclip}{\textsc{IS-clip}\xspace}
\newcommand{\MDR}{\textsc{M-DR}\xspace}

\newcommand{\DR}{\textsc{DR}\xspace}
\newcommand{\DM}{\textsc{DM}\xspace}

\newcommand{\Q}{\mathbb{Q}}
\newcommand{\Real}{\mathbb{R}}
\newcommand{\Borel}{\mathbb{B}}

\newcommand{\F}{\mathcal F}

\newcommand{\A}{\mathcal A}

\newcommand{\X}{\mathcal X}

\renewcommand{\L}{\mathcal L}

\newcommand{\M}{\mathcal M}

\newcommand{\E}{\mathbb E}

\newcommand{\Prob}{\mathbb P}
\newcommand{\Ind}{\mathbbm{1}}

\newcommand{\R}{\mathfrak{R}}

\newcommand{\CVaR}{\textnormal{CVaR}}

\newcommand{\varXA}{\sigma^2}

\newcommand{\Var}{\mathbb{V}}

\newcommand{\wt}{\widetilde}
\newcommand{\wh}{\widehat}
\newcommand{\wb}{\overline}
 
\newcommand{\cvara}{\text{CVaR}_\alpha}

\newtheorem{theorem}{Theorem}[section]
\newtheorem{lemma}{Lemma}[section]
\newtheorem{remark}{Remark}[section]
\newtheorem{corollary}{Corollary}[section]

\newtheorem{definition}{Definition}[section]

\newcommand{\leqi}[1]{\textcolor{magenta}{LL:#1}}

\title{Off-Policy Risk Assessment in Contextual Bandits}

\author[1]{Audrey Huang}
\author[1]{Liu Leqi}
\author[1]{Zachary C. Lipton}
\author[2]{Kamyar Azizzadenesheli}
\affil[ ]{\texttt{\href{mailto:audreyh@andrew.cmu.edu}{\textcolor{black}{audreyh@andrew.cmu.edu}},\href{mailto:leqil@cs.cmu.com}{\textcolor{black}{leqil@cs.cmu.edu}},\href{mailto:zlipton@cmu.edu}{\textcolor{black}{zlipton@cmu.edu}},\href{mailto:kamyar@purdue.edu}{\textcolor{black}{kamyar@purdue.edu}}}}
\affil[1]{Machine Learning Department, Carnegie Mellon University}
\affil[2]{Department of Computer Science, Purdue University}

\begin{document}

\maketitle

\begin{abstract}
  Even when unable to run experiments,
practitioners can evaluate prospective policies, 
using previously logged data.
However, while the bandits literature
has adopted a diverse set of objectives, 
most research on \emph{off-policy evaluation}
to date focuses on the expected reward. 
In this paper, we introduce 
Lipschitz risk functionals, 
a broad class of objectives
that subsumes conditional value-at-risk (CVaR),
variance, mean-variance, 
many distorted risks, and CPT risks, among others.
We propose \emph{Off-Policy Risk Assessment} (OPRA),
a framework that first estimates a target policy's CDF
and then generates plugin estimates 
for any collection of Lipschitz risks,
providing finite sample guarantees
that hold simultaneously 
over the entire class. 
We instantiate OPRA with 
both importance sampling 
and doubly robust estimators.
Our primary theoretical contributions are
(i) the first uniform concentration inequalities 
for both CDF estimators in contextual bandits
and (ii) error bounds on our Lipschitz risk estimates, 
which all converge at a rate of $O(1/\sqrt{n})$.
\end{abstract}

\section{Introduction}
\label{sec:introduction}
Many practical tasks, 
including medical treatment \citep{tewari2017ads}
and content recommendation \citep{li2010contextual}
are commonly modeled
within the contextual bandits framework.
In the online setting, an agent
observes a context at each step
and chooses among the available actions.
The agent then receives 
a context-dependent reward 
corresponding to the action taken,
but cannot observe the rewards
corresponding to alternative actions. 
In a healthcare setting, 
the observed context might be a vector 
capturing vital signs, lab tests,
and other available data,
while the action space might consist
of the available treatments.
The reward to optimize could 
be a measure of patient health
or treatment response.

While contextual bandits research 
has traditionally focused 
on the expected reward,
stakeholders often care 
about other risk functionals
(parameters of the reward distribution)
that express real-world desiderata
or have desirable statistical properties.
For example, investors assess mutual funds
via the Sharpe ratio,
which normalizes returns 
by their variance \citep{sharpe1966mutual}.
Related works in reinforcement learning (RL)
have sought to estimate 
the variance of returns 
\citep{sani2013risk, tamar2016learning} 
and to optimize the mean return
under variance constraints \citep{mannor2011mean}.
In safety-critical and financial applications,
researchers often measure
the conditional value-at-risk (CVaR),
which captures the expected return 
among the lower $\alpha$ quantile of outcomes
\citep{rockafellar2000optimization, keramati2020being}. 
In an emerging line of RL works,
researchers have explored
other risk functionals,
including cumulative prospect weighting~\cite{gopalan2017weighted}, 
distortion risk measures~\cite{dabney2018implicit},
and exponential utility functions~\cite{denardo2007risk}.

In many real-world problems otherwise suited 
to the contextual bandits framework,
experimentation turns out to be 
prohibitively expensive or unethical.
In such settings, 
we might hope
to evaluate prospective policies
using the data collected
under a previous policy.
Formally, this problem is called 
\emph{off-policy evaluation},
and our goal is to evaluate 
the performance of a target policy $\pi$
using data collected 
under a behavior policy $\beta$.
While most existing research
focuses on estimating 
the expected value of the returns
\citep{dudik2011doubly, dudik2014doubly},
one recent paper evaluates
the variance of returns
\citep{chandak2021highconfidence}.

In this paper, we propose 
practical methods and 
the first sample complexity guarantees
for \emph{off-policy risk evaluation},
addressing a diverse set of objectives
of interest to researchers and practitioners.
Towards this end, 
we introduce \emph{Lipschitz risk functionals}
which encompass all objectives 
for which the risk 
(i) depends only on the CDF of rewards;
and (ii) is Lipschitz with respect
to changes in the CDF 
(as assessed via the sup norm).
We prove that for bounded rewards, 
this class subsumes many risk functionals
of practical interest, including variance, 
mean-variance, conditional value-at-risk,
and cumulative prospect weighting, among others.

Thus, given accurate estimates
of the CDF of rewards under $\pi$,
we can accurately estimate Lipschitz risks.
Moreover, (sup norm) error bounds on our CDF estimates 
imply error bounds on the corresponding plugin estimates
for any Lipschitz risks.
The key remaining step is to establish 
finite sample guarantees on the error 
in estimating the target policy's 
CDF of rewards.
Our analysis centers on 
an importance sampling estimator 
(Section \ref{sec:is_estimator}), 
and a variance-reduced 
doubly robust estimator 
(Section \ref{sec:dr_estimator}).
We derive finite sample concentrations 
for both CDF estimators, 
showing that they achieve 
the desired $O(1/\sqrt{n})$ rates,
where $n$ is the sample size.
Moreover, the estimation error 
for any Lipschitz risk 
is scales with its Lipschitz constant, 
and similarly converges as $O(1/\sqrt{n})$.

We assemble these results
into an algorithm called OPRA
(Algorithm~\ref{algo:opre_general}) 
that outputs a comprehensive risk assessment 
for a target policy $\pi$, 
using any set of Lipschitz risk functionals.
Notably, because all risk estimates
share the same underlying CDF estimate,
our error guarantees hold simultaneously 
for all estimated risk functionals in the set, 
regardless of the cardinality (Section~\ref{sec:risk}). 
Finally, 
we present experiments that demonstrate the practical applicability our estimators.

\section{Related Work}

The study of risk functionals and risk-aware algorithms 
is core to the decision making literature \cite{artzner1999coherent,rockafellar2000optimization,krokhmal2007higher,shapiro2014lectures,acerbi2002spectral,prashanth2016cumulative,jie2018stochastic}. 
In supervised learning, \cite{soma2020statistical,curi2019adaptive} 
study generalization properties under the CVaR risk functional, 
while \cite{lee2020learning} study a variety of risk functionals.
\cite{khim2020uniform} considers 
the uniform convergence of $L$ risks
that are induced by CDF-dependent weighting functions 
and generalize CVaR and cumulative prospect theory (CPT)
inspired risks~\cite{liu2019human}.
Uniform convergence of worst-case risks
defined by $f$-divergences, 
which is a generalization of CVaR, 
is studied in \cite{duchi2018learning}. 
A recent line of research on developing 
the pointwise concentration of CVaR 
can be found in~\cite{brown2007large,thomas2019concentration,prashanth2020concentration}.

In the bandit literature, 
many works address regret minimization problems
using risk functionals; 
popular examples include the CVaR, 
value-at-risk, and mean-variance
\cite{casselgeneral,sani2013risk,vakili2015mean,zimin2014generalized}.
\cite{tamkin2019distributionally} studies
optimistic UCB exploration for optimizing CVaR
while \cite{chang2020risk,baudry2020thompson} 
study Thompson sampling,
and \cite{kagrecha2019distribution,bubeck2011x} 
study regret minimization for 
linear combinations of the mean and CVaR.
Using the CPT risk functional, 
\cite{gopalan2017weighted} considers regret minimization 
in both $K$-armed bandits and linear contextual bandits.
\cite{torossian2019mathcal,munos2014bandits} 
tackle the problem 
of black-box function optimization 
under different risk functionals. 

In off-policy evaluation,
we face an additional challenge 
due to the discrepancy 
between the data distribution
and that induced by the target policy.
Importance sampling (\IS) estimators 
are among the most prominent methods 
for dealing with distribution shift \citep{Aleksandrov68,horvitz1952generalization, shimodaira2000improving}. 
Doubly robust (\DR) estimators 
\citep{robins1995semiparametric, bang2005doubly} 
leverage (possibly misspecified) models
to achieve lower variance 
without sacrificing consistency.
These estimators have been adapted 
for off-policy evaluation in multi-armed bandits \citep{li2015toward,thomas2015high,chandak2021highconfidence}, 
contextual bandits \citep{dudik2011doubly,dudik2014doubly,wang2017optimal}, 
and Markov decision processes
\citep{jiang2016doubly,thomas2016data}.

In addition, distributional reinforcement learning methods
have gained traction in recent years. 
These methods, introduced by \cite{bellemare2017distributional},
work with the full distribution of returns 
and were subsequently improved upon by
\cite{dabney2018implicit,dabney2018distributional}. 
Notably, \cite{dabney2018implicit} 
uses the learned distribution 
to optimize a number of risk functionals,
including the CVaR and other distorted risk functionals. 
\cite{keramati2020being} leverages a similar method 
to learn the distribution of returns,
but uses optimistic distribution-dependent exploration 
to optimize the CVaR in MDPs. 
Along similar lines, \cite{tamkin2019distributionally} 
considers CVaR regret minimization using UCB exploration 
in the multi-armed bandit setting, 
and uses an empirical estimate 
of the reward distribution for each arm 
in order to evaluate the CVaR.

For empirical CDF estimation, 
the seminal work of \cite{dvoretzky1956asymptotic} 
provides an approximation-theoretic 
concentration bound 
which was later tightened by \cite{massart1990tight}. 
\cite{alexander1984probability} provides concentration bounds 
on probability estimates of arbitrary but structured measurable sets. 
Later, the works of \cite{vapnik2006estimation,ganssler1979empirical} 
systematically improved Alexander's inequality. 
For more details on concentrations of probability estimates,
we refer readers to \cite{devroye2013probabilistic}.

After deriving our key results, 
we learned of a prior independent
(but then unpublished) work 
\citep{chandak2021universal} 
that also employs importance sampling
to estimate CDFs for the purpose 
of providing off-policy estimates 
for parameters of the reward distribution.
However, they do not establish uniform concentration
of their estimates or formally relate 
the parameter and CDF errors,
leaving open questions concerning the convergence
(both asymptotically and in finite samples)
of the parameter estimates.
Our work formulates both importance sampling 
and variance-reduced doubly robust estimators
and provides the first uniform 
finite sample concentration bounds 
for both types of CDF and risk estimates.

\section{Problem Setting}\label{sec:problem_setting}
 
We denote contexts by $X$
and the corresponding context space by $\X$.  
Similarly, we denote actions by $A$ 
and the corresponding action space by $\A$. 
We study the contextual bandit problem characterized 
by a fixed probability measure over context space $\X$,
and a reward function that maps from tuples 
of contexts and actions to rewards:
$\mathcal{R}: \X \times \A \rightarrow \Real$. 
In the off-policy setting, 
we have access to a dataset $\mathcal{D}$
generated using a behavior policy $\beta$ 
that interacts with 
the environment for $n$ rounds as follows: 
at each round, a new context $X$ is drawn
and then the policy $\beta$ 
chooses an action $A \sim \beta(\cdot|X)$.
The environment then reveals the reward 
$R \sim \mathcal{R}(\cdot|X,A)$ 
for only the chosen action $A$. 
Running this process for $n$ steps
generates a dataset
$\mathcal{D} := \lbrace x_i, a_i, r_i \rbrace_{i=1}^n$. 
In the off-policy evaluation setting, 
our goal is to evaluate the performance 
of a target policy $\pi$, 
using only a dataset $\mathcal{D}$.

Next, we can express our sample space
in terms of the contexts, actions, and rewards:
$\Omega = \left(\mathcal{X} \times \mathcal{A} \times \Real \right)$.
Let $(\Omega, \F, \Prob_\beta)$ be the probability space 
induced by the behavior policy $\beta$, 
and $(\Omega, \F, \Prob)$ the probability space 
induced by the target policy $\pi$. 
We assume that $\Prob$ is absolutely continuous 
with respect to $\Prob_\beta$. 
For any context $x$ and action $a$,
the importance weight expresses the ratio
between the two densities 
$\weight(\omega)= \weight(a,x) = \frac{\beta(a|x)}{\pi(a|x)}$, 
and the maximum weight $w_{\max} = \sup_{a,x} w(a,x)$ is simply the supremum 
taken over all contexts and actions. Further, let $w_\mathbbm{2} = \E_{\Prob_\beta}\left[\weight(A,X)^2\right]$ denote the exponential 
of the second order R\'enyi divergence. 
Note that by definition,
$w_\mathbbm{2} \leq w_{\max}$,  
and in practice, we often have 
$w_\mathbbm{2} \ll w_{\max}$.

Finally, we introduce 
some notation
for describing \CDF{}s:
For any $t\in\Real$,
let $F(t) = \E_{\Prob}[\Ind_{\{R \leq t\}}]$ 
denote the \CDF under the target policy; 
further, let $G(t;X,A) = \E_{\Prob}[\Ind_{\{R \leq t\}} \vert X,A] = \E_{\Prob_\beta}[\Ind_{\{R \leq t\}} \vert X,A]$ denote the \CDF of rewards 
conditioned on a context $X$ and action $A$, which is independent of the policy. 
Lastly, for any $t\in\Real$, 
we denote the variance by  
$\varXA(t;X,A) = \Var_\Prob\left[\Ind_{\lbrace R\leq t\rbrace}\vert X,A\right] = \Var_{\Prob_\beta}\left[\Ind_{\lbrace R\leq t\rbrace}\vert X,A\right]$.

\section{Lipschitz Risk Functionals}\label{sec:risk_functionals}
We now introduce Lipschitz risk functionals,
a novel class of objectives 
for which absolute differences in the risk 
are bounded by sup norm differences in the CDF of rewards. 
After formally defining the class, 
we provide an in-depth review 
of common risk functionals 
and their relationship 
to the \CDF of rewards. 
When possible, we derive 
the associated Lipschitz constants,
when rewards are bounded on support $[0, D]$,
relegating all proofs to Appendix~\ref{appendix:lipschitz_risk}.

\subsection{Defining the Lipschitz Risk Functionals}
\label{sec:lipschitz-bounded-rv}
The Lipschitz risk functionals are a subset 
of the broader family of 
\emph{law-invariant risk functionals}.
Formally, let $Z\in \L_\infty(\Omega, \F_Z,\Prob_Z)$ 
denote a real-valued random variable that admits a \CDF 
$F_Z\in \L_\infty(\Real, \Borel(\Real))$. 
A \emph{risk functional} $\rho$
is a mapping from a space of random variables to the space of real numbers
$\rho: \L_\infty(\Omega, \F_Z,\Prob_Z) \rightarrow \Real$.
Any risk functional $\rho$
is said to be law-invariant if $\rho(Z)$ depends 
only on the distribution of $Z$~\citep{kusuoka2001law}.  

\begin{definition}[Law-Invariant Risk Functional]
A risk functional
$\rho:\L_\infty(\Omega, \F,\Prob)  \rightarrow \Real$, 
is law-invariant
if for any pair of random variables $Z$ and $Z'$,
$
    F_Z = F_{Z'}\;\Longrightarrow\; \rho(Z)=\rho(Z'). 
$
\end{definition}

When clear from the context,
we sometimes abuse notation
by writing $\rho(F_Z)$ 
in place of $\rho(Z)$. 
In general, it may not be practical 
to estimate risk functionals 
that are not law invariant
from data~\citep{balbas2009properties}.
Thus focusing on law-invariant risks 
is only mildly restrictive.

We can now formally define 
the Lipschitz risk functionals:
\begin{definition}[Lipschitz Risk Functional]
A law invariant risk functional $\rho$ is $L$-Lipschitz 
if for any pair of \CDF{}s $F_Z$ and $F_{Z'}$ and some $L \in (0, \infty)$, 
it satisfies 
\begin{align*}
    |\rho(F_Z)-\rho(F_{Z'})|\leq L \|F_Z-F_{Z'}\|_\infty.  
\end{align*}
\end{definition}
A risk functional is $L$-Lipschitz if, 
for any two random variables $Z, Z'$, 
its value is upper bounded
by the sup-norm of the difference between
their corresponding CDFs.
The significance of this Lipschitzness property 
in the contextual bandit setting is that,
given a high confidence bound on the error 
of the estimated CDF of rewards for a policy $\pi$, 
we can obtain a high confidence bound on its evaluation 
under any $L$-Lipschitz law-invariant risk functional
on the distribution of rewards.

\subsection{Overview of Common Risk Functionals (and their Lipschitzness)}
We now briefly describe some popular classes
of risk functionals and their axiomatic definitions. When possible, we
derive their associated Lipschitz constants. 

First, we enumerate
a set of prominent axioms
explored in the current literature
\citep{artzner1999coherent,sereda2010distortion}. 
Consider a pair of random variables $Z$ and $Z'$, 
we have the following axioms:

\begin{enumerate}
    \item \label{Monotonicity} Monotonicity: $\rho(Z) \leq \rho(Z')$ whenever $Z \leq Z'$.
    \item \label{Subadditivity} Subadditivity: $\rho(Z+Z') \leq \rho(Z) + \rho(Z')$.
    \item \label{Additivity} Additivity: $\rho(Z+Z') =\rho(Z) + \rho(Z')$ if $Z$ and $Z'$ are co-monotonic random variables 
    (i.e., there exists a random variable $Y$ 
    and weakly increasing functions $f, g$ 
    such that $Z = f(Y)$ and $Z' = g(Y)$). 
    \item \label{Translation_invariance}Translation invariance:
    $\rho(Z + c) = \rho(Z) + c, \forall c \in \Real$.
    \item \label{Positive_homogeneity}Positive homogeneity: 
    $\rho(tZ) = t\rho(Z)$ for $t > 0$.
    \item \label{Bounded_above} Bounded above by the maximum cost, 
    i.e., $\rho(Z)\leq \max(Z)$. 
    \item \label{Bounded_below} Bounded below by the mean cost, 
    i.e., $\rho(Z)\geq \E[Z]$. 
\end{enumerate}

From this set of axioms,
one can define a class of risk functionals
by choosing the subset best suited
to the problem at hand.

\paragraph{Coherent Risk Functionals.} 
The set of risk functionals that satisfy 
monotonicity~(Axiom \ref{Monotonicity}),
subadditivity~(Axiom \ref{Subadditivity}),
translation invariance~(Axiom \ref{Translation_invariance}), 
and positive homogeneity~(Axiom \ref{Positive_homogeneity}), and positive homogeneity 
(see Appendix~\ref{appendix:lipschitz_risk}),  
constitute the \emph{coherent risk functionals} \citep{artzner1999coherent,delbaen2002coherent}. Further, if a law-invariant coherent risk functional 
additionally satisfies Additivity (Axiom \ref{Additivity}),
it is said to be a \emph{spectral risk functional}~\citep{jouini2006law,acerbi2002spectral}.

While not all coherent risk functionals 
are law-invariant, 
nearly all of those commonly 
addressed in the literature are.
Examples include expected value, 
conditional value-at-risk (CVaR), 
entropic value-at-risk,
and mean semideviation~\citep{chang2020risk, tamkin2019distributionally, tamar2015policy, shapiro2014lectures}. 
Others include the Wang transform function~\cite{wang1996premium} 
and the proportional hazard (PH) 
risk functional~\cite{wirch2001distortion}.

\paragraph{Distorted Risk Functionals.}
When the random variable $Z$ 
is required to be non-negative, 
law-invariant coherent risk functionals 
are examples of the more general class 
of law-invariant \emph{distorted risk functionals}
\citep{denneberg1990distorted,wang1996premium,wang1997axiomatic,balbas2009properties}. 
For $Z \geq 0$, a distorted risk functional has the following form 
\begin{align*}
    \rho(F_Z) = \int_0^\infty g(1-F_Z(t))dt,  
\end{align*}
where the distortion function $g:[0,1]\rightarrow[0,1]$ 
is an increasing function with $g(0)=0$ and $g(1)=1$. 
Distorted risk functionals are coherent 
if and only if $g$ is concave \citep{wirch2001distortion}. 
For example, when $g(s) = \min\{\frac{s}{1-\alpha},1\}$ 
for $s\in[0,1]$ and $\alpha \in (0, 1)$,
$\CVaR$ at level $\alpha$ is recovered. 
When $g$ is the identity map, 
the distorted risk functional 
is the expected value. 
The Wang risk functional at level $\alpha$~\citep{wang1996premium} 
is recovered when $g(s)=F(F^{-1}(s)-F^{-1}(\alpha))$,
and the proportional hazard risk functional 
can by obtained by setting $g(s)=s^\alpha$ for $\alpha < 1$. 
Not all distorted risk functionals are coherent.
For example, setting $g(s)=\Ind_{\{s\geq 1-\alpha\}}$
recovers the value-at-risk (VaR),
which is not coherent. 

Distorted risk functionals have many desirable theoretical properties.
They are translation invariant (Axiom \ref{Translation_invariance})
and positive homogeneous~(Axiom \ref{Positive_homogeneity}), 
and are defined utilizing~(Axiom \ref{Bounded_above}) 
and~(Axiom \ref{Bounded_below})~\citep{wirch2001distortion}.
They satisfy~Axiom \ref{Bounded_below} 
if and only if $g(s)\geq s~\forall s\in[0,1]$~\citep{wirch2001distortion}, 
and are subadditive~(Axiom \ref{Subadditivity}) 
if and only if $g$ is concave, 
which preserves second order stochastic dominance~\citep{wang1996premium}. 
In addition, all distorted risk functionals 
preserve stochastic first order dominance~\citep{wirch2001distortion}.

\begin{lemma}[Lipschitzness of Coherent and Distorted Risk Functionals]\label{lem:distorted_lipschitz}
On the space of random variables with support in $[0,D]$, 
the distorted risk functional of any $\frac{L}{D}$-Lipschitz distortion function $g:[0,1]\rightarrow[0,1]$, i.e., $|g(t)-g(t')|\leq \frac{L}{D}|t-t'|$,  
is a $L$-Lipschitz risk functional.
\end{lemma}

\begin{remark}[Expected Value and CVaR]
 Both expected value and CVaR are examples of distorted risk functionals. 
 Then using Lemma~\ref{lem:distorted_lipschitz}, 
 on the space of random variables with support in $[0,D]$, 
 the expected value risk functional is $D$-Lipschitz 
 because $g$ is the identity and thus 1-Lipschitz. 
 On the same space, the risk functional $\cvara$ 
 is $\frac{D}{\alpha}$-Lipschitz 
 because $g$ is $\frac{1}{\alpha}$-Lipschitz. 
\end{remark}

\paragraph{Cumulative Prospect Theory (CPT) Risk Functionals.} 
CPT risks~\cite{prashanth2016cumulative} take the form:
\begin{align*}
    \rho(F_Z) = \int_0^{+\infty}g^+\left(1 - F_{u^+(Z)}(t)
    \right)dt - \int_0^{+\infty}g^-\left(1 - F_{u^-(Z)}(t)\right) dt, 
\end{align*}
where $g^+, g^-$: $[0, 1] \rightarrow [0, 1]$, 
$g^{+/-}(0) = 0$, 
and $g^{+/-}(1) = 1$.
The functions $u^+, u^-: \Real \rightarrow \Real_+$ are continuous, 
with $u^+(z) = 0$ when $z \geq c$ and $u^-(z) = 0$ 
when $z < c$ for some constant $c \in \Real$.
Importantly, the CPT functional handles gains and losses separately. 
The functions $u^+, u^-$ compare the random variable $Z$
to a baseline $c$, 
and the distortion $g^+$ is applied to ``gains'' 
(when $Z \geq c$),
while $g^-$ is applied to ``losses'' (when $Z < c$).  

Note that the distortions $g^+, g^-$ may not necessarily be monotone functions. 
As a result, the distortion functionals can be seen as 
a special case of the CPT functional 
when $Z$ is nonnegative, 
$c = 0$, 
and $g$ is an increasing function.
Appropriate choices of $g^+, g^-$ 
can again be used to recover risk functionals such as the CVaR and VaR. 
We make note of the fact that, in the CPT literature, 
$g^{+/-}$ is chosen to necessarily have a fixed point 
where $g^{+/-}(s) = s$ for some $s \in (0, 1)$, 
although we do not make this assumption here.

In general, due to the general form of $g^{+/-}$ and the separate consideration of losses and gains, the CPT-inspired risk functional may not satisfy any of the defined axioms. However, additional assumptions on the distortions $g^+$ and $g^-$ may allow certain axioms to be satisfied. For example, if the random variable has nonnegative support and the threshold $c$ is set to be 0 so that only gains are observed, and $g^+$ is additionally increasing, we recover the distorted risk functionals with axioms specified above. If $g^+$ is additionally concave, then we recover the coherent risk functionals. 

\begin{lemma}[Lipschitzness of CPT Functional]\label{lem:cpt_lipschitz}
On the space of random variables with support in $[0,D]$, 
if the CPT distortion functions $g^+$ and $g^-$
are both $\frac{L}{D}$-Lipschitz, 
then the CPT risk functionals is $L$-Lipschitz. 
\end{lemma}

\paragraph{Other Risk Functionals.} 
The variance, mean-variance,
and many other popular risks 
do not fit easily into 
the aforementioned classes, 
but are nevertheless law-invariant.
For example, for a nonnegative random variable $Z$, 
the variance is defined as 
$\rho(F_Z) = 2\int_{0}^\infty t(1-F_Z(t))dt - \left(\int_{0}^\infty (1-F_Z(t))dt \right)^2.$
Moreover, the variance and mean-variance 
are both $L$-Lipschitz. 

\begin{lemma}[Lipschitzness of Variance]\label{lemma:variance_lipschitz}
 On the space of random variables with support in $[0,D]$,
 variance is a $3D^2$-Lipschitz risk functional.
\end{lemma}

A number of recent papers 
have addressed risk functionals 
expressed as weighted combinations 
of others, e.g., mean-variance~\cite{sani2013risk}.
Other papers have optimized constrained objectives, 
such as expected reward constrained by variance 
or CVaR below a certain threshold 
\cite{chow2017risk, prashanth2018risk}.
When expressed as Lagrangians, 
these objectives can also be expressed 
as weighted combinations 
of the risk functionals involved. 
We extend the Lipschitzness property 
to risk functionals of this form: 
\begin{lemma}[Lipschitzness of Weighted Sum of Risk Functionals]\label{lem:weighted_lipschitz}
Let $\rho$ be a weighted sum 
of risk functionals $\rho_1,...,\rho_K$ 
that are $L_1,...,L_K$-Lipschitz, respectively, 
with weights $\lambda_1,....,\lambda_K > 0$, i.e.,  
$\rho(Z) = \sum_{k=1}^K \lambda_k \rho_k(Z)$.
Then $\rho$ is $\sum_k \lambda_k L_k $-Lipschitz. 
\end{lemma}
\begin{remark}
Note that mean-variance is given 
by $\rho(Z) = \E[Z] + \lambda \Var(Z)$ 
for some $\lambda > 0$. 
Then, using Lemma \ref{lem:weighted_lipschitz}, 
we immediately obtain that mean-variance 
is $(1 + 3\lambda D^2)$-Lipschitz 
for bounded random variables.
\end{remark}

Though we have provided many examples of Lipschitz risk functionals in this section, it is worth noting that there are a number of risk functionals that do not satisfy the Lipschitzness property, such as the value-at-risk (VaR).
For the sake of brevity, we omit consideration of such risk functionals in this paper, and outline future avenues of research on this topic in the discussion.

\section{Off-Policy CDF Estimation}\label{sec:cdf_estimator} 
This section describes our method for 
high-confidence off-policy estimation of $F$, 
the CDF of returns under the policy $\pi$. 
The key challenge in estimating $F$
is that the reward samples are observed 
only for actions taken by the behavior policy $\beta$.  
To overcome this limitation, 
one intuitive solution is to reweight
the observed samples according to 
their importance sampling (IS) weight (Section~\ref{sec:is_estimator}).
However, IS estimators are known to suffer from high variance.
To mitigate this, we define the first doubly robust CDF estimator (Section~\ref{sec:dr_estimator}). 

\subsection{CDF Estimation with Importance Sampling (IS)}\label{sec:is_estimator}
Given an off-policy dataset $\mathcal{D} = \lbrace x_i, a_i, r_i \rbrace_{i=1}^n$, 
we define the following nonparametric IS-based estimator for the empirical CDF, 
\begin{equation}\label{is_estimator}
    \widehat{F}_{\text{IS}}(t) := \frac{1}{n}\sum_{i=1}^n w\left(a_i, x_i\right)\mathbbm{1}_{\{r_i \leq t\}},
\end{equation}
where $w(a, x) = \frac{\pi(a|x)}{\beta(a|x)}$ are the importance weights. 
The IS estimator is pointwise-unbiased, 
with variance given below:

\begin{lemma}\label{lem:is_bias_var}
    The IS estimator~\eqref{is_estimator} is unbiased and its variance is 
    \begin{align*}
    \Var_{\Prob_\beta}\left[\wh{F}_{\text{IS}}(t)\right] &= \frac{1}{n}\E_{\Prob_\beta}\left[w(A,X)^2 \sigma^2(t;X,A) \right] + \frac{1}{n}\Var_{\Prob_\beta}\left[\E_{\Prob_\beta}\left[w(A,X) G(t;X,A)\vert X\right]\right] \\ 
    &\quad\quad+ \frac{1}{n}\E_{\Prob_\beta}\left[\Var_{\Prob_\beta}\left[w(A,X) G(t;X,A)\vert X\right]\right]
\end{align*}
\end{lemma}

The expression for variance is broken down into three terms. 
The first term represents randomness in the rewards.
The second term represents variance 
due to the randomness over contexts $X$.
The final term is the penalty arising 
from using importance sampling, 
and is proportional to the importance sampling weights $\weight$ 
and the true CDF of conditional rewards $G$. 
The variance contributed by the third term
can be large
when the weights $\weight$
have a wide range, which occurs
when $\beta$ assigns extremely small probabilities
to actions where $\pi$ assigns high probability.   

Due to the use of importance sampling weights, 
the estimated CDF $\widehat{F}_{\text{IS}}(t)$ 
may be greater than 1 for some $t$,
even though a valid \CDF must be in the interval $[0, 1]$ for all $t$.
To mitigate this problem, a weighted importance sampling (\WIS) estimator can be used,
which normalizes each importance weight by the sum of importance weights: 
\begin{align*}
    \wh{F}_\WIS(t) = \frac{1}{\sum_{j=1}^n w(a_j, x_j) } \sum_{i=1}^n w(a_i, x_i) \mathbbm{1}_{\{r_i \leq t\}},
\end{align*}
which \cite{chandak2021universal} shows is a biased but uniformly consistent estimator. Another option is the clipped estimator IS-Clip~\eqref{is_clip_estimator}, which simply limits the estimator to the unit interval: 
\begin{equation}\label{is_clip_estimator}
    \widehat{F}_{\ISclip}(t) := \min\{\widehat{F}_{\IS}(t), 1\}
\end{equation}
Although $\widehat{F}_{\ISclip}$ 
has lower variance than the IS estimator, it is potentially biased.

However, given finite samples, 
we can bound with high confidence the sup-norm error between $\wh{F}_\ISclip$ and $F$, 
in Theorem \ref{thm:is-cdf} below (proof in Appendix~\ref{proof:is-error}): 

\begin{theorem}\label{thm:is-cdf}
Given $n$ samples drawn from $\Prob_\beta$, 
for the \IS estimator $\wh F_{\IS}(t)$, we have
\begin{equation}\label{eq:IS-concentration}
    \Prob_\beta\left(\Vert \wh{ F}_{\ISclip} -  F\Vert_{\infty} \leq \varepsilon_{\IS_1}:= \sqrt{\frac{8\weight^2_{\max}}{n}\log(4/\delta)}\right)\geq  1-\delta.
\end{equation}
or, based on $w_{\mathbbm{2}}$, we obtain a Bernstein-style bound, 
\begin{align}\label{eq:IS-concentration-Bernstein}
    \Prob_\beta\left(\left\Vert\wh{ F}_{\ISclip} -  F \right\Vert_\infty  \leq \varepsilon_{\text{IS}_2}:= \frac{4\weight_{\max}\log(4/\delta)}{n}+2\sqrt{\frac{2w_{\mathbbm{2}}\log(4/\delta)}{n}}\right)\geq 1-\delta
\end{align}
\end{theorem}

When $w_{\mathbbm{2}} \ll w_{max}$, we observe that inequality~\eqref{eq:IS-concentration-Bernstein} is more favorable than  inequality~\eqref{eq:IS-concentration}. 
Theorem~\ref{thm:is-cdf} demonstrates that the $\wh{F}_\ISclip$ uniformly converges to the true \CDF at a rate of $O(1/\sqrt{n})$, with the uniform consistency of $\wh{F}_\ISclip$ as an immediate consequence.
To the best of our knowledge, it is the first 
DKW-style concentration inequality 
on the importance sampling estimator \CDF estimator
in off-policy evaluation. 
The bound has explicit constants 
and subsumes the classical DKW inequality.

\subsection{Model-Based CDF Estimation}\label{sec:dm_estimation}
As we have shown previously, IS estimators 
can suffer from high variance, which can be limiting in practice. 
However, in many practical applications, we may have access to
a model $\overline{G}(t;X,A)$ of the conditional distribution $G(t;X,A)$, 
which can be used in estimation with very low variance. 
In many cases, practitioners may have a model of $\overline{G}$ from expert studies or from a simulator, 
or can 
form a regression estimate of $\overline{G}$ from logged data.
One simple model-based estimator can then be obtained using the \emph{direct method}, which simply employs the model $\overline{G}$ for each observed context: 
\begin{equation}
    \wh{F}_\DM(t) = \frac{1}{n}\sum_{i=1}^n \overline{G}(t;x_i, \pi), \quad\text{where}\;\; \overline{G}(t;x_i, \pi) = \sum_a \pi(a|x_i) \overline{G}(t;x_i, a). 
\end{equation}
Because the DM estimator $\widehat{F}_\DM$ 
does not use importance weights, 
it can have significantly lower variance 
than the IS and DR estimators. 
In general, however, the DI estimator is biased, and 
its error flows directly from error in the model $\overline{G}$ 
(full derivations of the bias and variance
are given in Lemma \ref{lem:di_bias} 
of Appendix \ref{appendix:estimated_policy}). 
The magnitude and distribution of bias over 
the context and action space is difficult to  characterize.
In practice, $\overline{G}$ is often estimated
or modeled agnostic to the target policy,
and hence may not be well-approximated 
in areas that are important for $\pi$. 
If $\overline{G}$ is an accurate model 
of the conditional reward distribution, however, 
then $\widehat{F}_{\text{DM}}$ is a good approximation of $F$. 

\subsection{Doubly Robust (DR) CDF Estimation}\label{sec:dr_estimator}

We now define a {doubly robust (DR) CDF estimator} 
that takes advantage of both importance sampling and models $\overline{G}$ to obtain the best characteristics of both types of estimation. 
In particular, the DR estimator is unbiased, 
but has potentially significant reduction in variance. 
The DR estimator for the empirical CDF is defined to be 
\begin{equation}\label{dr_estimator}
    \widehat{F}_{\text{DR}}(t) := \frac{1}{n}\sum_{i=1}^n w(a_i, x_i) \Big(\mathbbm{1}_{\{r_i \leq t\}} - \overline{G}(t; x_i, a_i) \Big) + \overline{G}(t; x_i, \pi),
\end{equation}
where $\overline{G}(t;x,\pi) = \E_{\Prob_\beta}\left[\overline{G}(t;x,A) | x\right]$. Informally, the DR estimator 
takes the
model $\overline{G}$ as a baseline, 
using the available data 
to apply a correction.
While $\overline{G}$ alone may be biased,
the DR estimator is an unbiased estimator of $F$, 
and can have reduced variance
compared to the IS estimator: 
\begin{lemma}\label{lem:dr_bias_var}
    The DR estimator~\eqref{dr_estimator} is unbiased
    and its variance is
    \begin{align*}
    \Var_{\Prob_\beta}\left[ \wh{F}_{\DR}(t) \right] &= \frac{1}{n}\E_{\Prob_\beta}\left[ w(A,X)^2\sigma^2(t;X,A) \right] + \frac{1}{n}\Var_{\Prob_\beta}\left[\E_{\Prob_\beta}\left[ w(A,X)G(t;X,a) \vert X\right] \right] \\
    &\quad\quad+ \frac{1}{n}\E_{\Prob_\beta}\left[\Var_{\Prob_\beta}\left[ w(A,X)\left(G(t;X,A) - \overline{G}(t;X,A)\right) \vert X \right]\right]
\end{align*}
\end{lemma}

The variance reduction advantage of the DR estimator becomes apparent 
from a direct comparison of the three terms in the 
IS estimator variance (Lemma~\ref{lem:is_bias_var})  
and the DR estimator variance (Lemma~\ref{lem:dr_bias_var}). 
The first and second terms, which capture 
the variance in rewards and contexts, are identical. 
The third term, which represents the importance sampling penalty, is proportional to $G - \overline{G}$ in the DR estimator, but proportional to $G$ in the IS estimator. 
When this difference $G - \overline{G}$ is smaller than $G$, 
which is often the case in practice, 
the third term has reduced variance in the DR estimator.  
The magnitude of variance reduction is greater
when the weights $\weight$ have a large range,  
which is precisely when large variance 
can become problematic in importance sampling.  

\begin{remark}[Double Robustness]
Although we consider the setting 
where the behavior policy $\beta$ is known, 
when the behavior policy is unknown and needs to be estimated, 
the estimator $\wh F_\DR$ is consistent
when either $\overline{G}$ is consistent
or the policy estimator is consistent. 
This is where the name ``doubly robust" 
comes from. 
We demonstrate and discuss this fact further 
in Appendix \ref{appendix:estimated_policy}. 
\end{remark}

Although the DR estimator $\widehat{F}_{\text{DR}}$ has 
desirable reductions in variance, 
given finite samples, 
it is not guaranteed to be a valid \CDF. 
Like the IS estimator,  
the DR estimator may be greater than $1$ for some $t$
due to the use of importance weighting. 
However, it may also be negative at some $t$
as a consequence of the subtracted term in~\eqref{dr_estimator}.  
As an additional consequence of this term, 
the DR estimator is not guaranteed to be a monotone function. 
As a result, 
in order to use the DR CDF estimate for risk estimation, 
we must transform $\widehat{F}_\DR$ 
into a monotone function bounded in $[0, 1]$. 
Examples of such transformations include isotonic approximation~\citep{smith1987best} 
and monotone $\L_p$ approximation~\citep{darst1985best}. 

For our analysis, however, we consider a simple monotone transformation 
that involves an accumulation function, 
which does not allow the CDF to decrease, 
followed by a clipping to $[0, 1]$: 
\begin{equation}\label{eq:dr_monotone} 
\widehat{F}_\MDR(t) = \text{Clip}\left\lbrace \max_{t' \leq t}\wh{F}_\DR(t'), 0, 1 \right\rbrace,
\end{equation}
which is a uniformly consistent estimator, as the following concentration guarantee shows: 
\begin{theorem}\label{thm:dr_error}
The monotone transformation of the DR estimator $\widehat{F}_\MDR(t)$ satisfies 
\begin{equation}
    \Prob_\beta\left( 
    \left\Vert \wh{F}_\MDR - F \right\Vert_{\infty} \leq \epsilon_\DR := \sqrt{\frac{72w_{\max}^2}{n}\log\left(\frac{8n^{1/2}}{\delta}\right)} \right) \geq 1-\delta. 
\label{eq:dr-agnostic-bound}
\end{equation}
\end{theorem}

The purpose of Theorem \ref{thm:dr_error} 
is to show the dependence of the error 
on the importance weights $w_{max}$ 
and on the finite sample size $n$. 
Using the \MDR estimator, 
we again recover a sample complexity 
of $\wt{O}\left(1/\sqrt{n}\right)$. 
The proof is given in Appendix \ref{appendix:dr_error}. 
Note that~\eqref{eq:dr-agnostic-bound}
does not depend on the error $\overline{G}-G$, 
which is the term 
responsible for variance reduction, 
as given in Lemma~\ref{lem:estimated_policy_dr}. 
A tighter bound for the DR estimator, 
which incorporates the error $G-\overline{G}$, 
remains an open problem, and we leave this to future work. 
We demonstrate empirically
in Section \ref{sec:experiments} that, in practice, 
we do achieve faster convergence and smaller empirical confidence intervals with 
the \MDR estimator.

\section{Off-Policy Risk Assessment}\label{sec:risk}
Given any law-invariant risk functional $\rho$ 
and 
CDF estimator $\wh{F}$, 
we can estimate the value of the risk functional as $\wh\rho:=\rho(\wh F)$.
However, the estimator $\wh\rho$ may be biased 
even if $\wh F$ is unbiased.  
For Lipschitz risk functionals introduced in Section~\ref{sec:risk_functionals}, 
we can obtain their
finite sample error bounds,
using the error bound of the CDF estimator. 
Further, a set of risk functionals of interest can be evaluated using the same estimated CDF, 
which suggests that the error bound of the CDF gives error bounds on the risk estimators that hold simultaneously. 

Theorem \ref{thm:confidence_general} 
utilizes 
our 
error bound
of the estimated CDF to derive error bounds 
for estimators of a set of Lipschitz risk functionals. 
As we showed in Section \ref{sec:risk_functionals}, 
most if not all commonly studied risk functionals 
satisfy the property of Lipschitzness, 
showing our result's wide applicability. 

\begin{theorem}\label{thm:confidence_general}
Given a set of Lipschitz risk functionals 
$\lbrace \rho_p \rbrace_{p=1}^P$ 
with Lipschitz constants $\lbrace L_p\rbrace_{p=1}^P$,
and a CDF estimator $\wh{F}$,
such that $\Vert \widehat{F} - F \Vert_\infty \leq \epsilon$ 
with probability at least $1-\delta$, 
we have with probability at least $1-\delta$ that for all $p \in \{1, \ldots P\}$,
\begin{equation*}\label{eq:confidence_general}
    \left\vert \rho_p(\wh{F}) - \rho_p(F) \right\vert \leq L_p\epsilon. 
\end{equation*}
\end{theorem}

Thus,
one powerful property of risk estimation 
using the estimated CDF approach 
is that, 
given a high-probability error bound on the CDF estimator, 
the corresponding error bounds on estimates of \emph{all} Lipschitz risk functionals of interest
hold \emph{simultaneously} with the same probability. 
Further, because the error of the IS
CDF estimator $\epsilon_{\text{IS}}$ (Theorem~\ref{thm:is-cdf}) and DR CDF estimator $\epsilon_\text{DR}$ (Theorem~\ref{thm:dr_error}) 
converge at a rate of $O(1/\sqrt{n})$,
Theorem~\ref{eq:confidence_general} shows that the error of all Lipschitz risk functional estimators
shrink at a rate of $O(1/\sqrt{n})$. 
Thus, $\rho_p(\wh F)$ are consistent risk functional estimators. 

Putting these results together, we now provide an algorithm, 
called OPRA (Algorithm \ref{algo:opre_general}),
which given an off-policy contextual bandit dataset 
and a set of Lipschitz risk functionals 
of interest, 
outputs for each risk functional an estimate of its value 
and a confidence bound. 
The algorithm first uses a valid CDF estimator, 
e.g., the clipped IS estimator~\eqref{is_clip_estimator} or monotonized DR estimator~\eqref{eq:dr_monotone}, 
to form $\wh{F}$ with sup-norm error $\epsilon$. 
OPRA then evaluates each $L_p$-Lipschitz risk functional $\rho_p$ on $\wh{F}$ to obtain $\wh\rho_p$, 
along with its upper and lower confidence bound 
$\wh\rho_p \pm L_p\epsilon$.

\begin{algorithm}[H]
\SetAlgoLined 
\KwIn{Dataset $\mathcal{D}$, policy $\pi$, probability $\delta$, models $\overline{G}$, 
Lipschitz risk functionals $\lbrace \rho_p\rbrace_{p=1}^P$ with Lipschitz constants $\lbrace L_p \rbrace_{p=1}^P$. }
Estimate the CDF using a valid \CDF estimator $\wh{F}$\; 
Compute the corresponding CDF estimation error $\epsilon$ such that $\mathbb{P}(\Vert F - \wh F\Vert_\infty < \epsilon) \geq 1-\delta$\; 
\For{p = 1 \ldots P}{
Estimate $\wh{\rho}_p = \rho_p(\wh{F})$\; 
}
\KwOut{Estimates with errors $\lbrace \wh{\rho}_p \pm L_p \epsilon\rbrace_{p=1}^P$.  }
\caption{Off-Policy Risk Assessment (OPRA)}
\label{algo:opre_general}
\end{algorithm}

OPRA can be used to obtain a full risk assessment of any given policy, using the input Lipschitz risk functionals of interest, 
which can include the popularly used mean, variance, and CVaR. 
As demonstrated in Theorem~\ref{thm:confidence_general}, 
the error guarantee on the risk estimators 
holds simultaneously for all $P$ risk functionals 
with probability at least $1-\delta$. 
Importantly, OPRA also demonstrates the computational efficiency 
of the distribution-centric risk estimation approach proposed in this paper. 
For a given $\pi$, the CDF only needs to be estimated once, 
and can be used repetitively 
to estimate the value of the risk functionals. 
Further, the error of the risk estimators are 
determined by the known error of the CDF estimator, 
multiplied by the known Lipschitz constants. 

\begin{remark}[Estimation of Risk Functionals That Are Not $L$-Lipschitz]
We have focused our discussion on the estimation 
of $L$-Lipschitz risk functionals 
due their generalizability and flexibility, 
and because we can characterize the rate at which the error decreases. 
Any law-invariant risk functional can actually be estimated using the CDF estimate, although the error or confidence of the estimate 
may have to be determined on a case-by-case basis
for each risk functional of interest. 
Further, the rate at which the error converges may not necessarily be known.
\end{remark}

\begin{remark}[Risk Functionals Estimation When Behavioral Policy Is Unknown]
    Although we work with known behavioral policy $\beta$ in this paper,
    previous works on off-policy evaluation have considered the case where the behavioral policy is unknown. 
    In such cases, an estimate or model of the policy, called $\wh\beta$, is instead used in CDF and risk estimation. 
    In Appendix~\ref{appendix:estimated_policy} we extend the bias, variance, error bound results to this setting. 
\end{remark}

\section{Empirical Studies}\label{sec:experiments}
In this section, we give empirical evidence 
for the effectiveness of the doubly robust (\DR) \CDF and risk estimates, 
in comparison to the importance sampling (\IS),
weighted importance sampling (\WIS), 
and direct method (\DM) estimates. 
Further, we demonstrate 
the convergence of the 
\CDF and risk estimation error 
in terms of the number of samples. 

\begin{figure}[t!]
    \centering
    \hspace*{-1cm}\includegraphics[width=\textwidth]{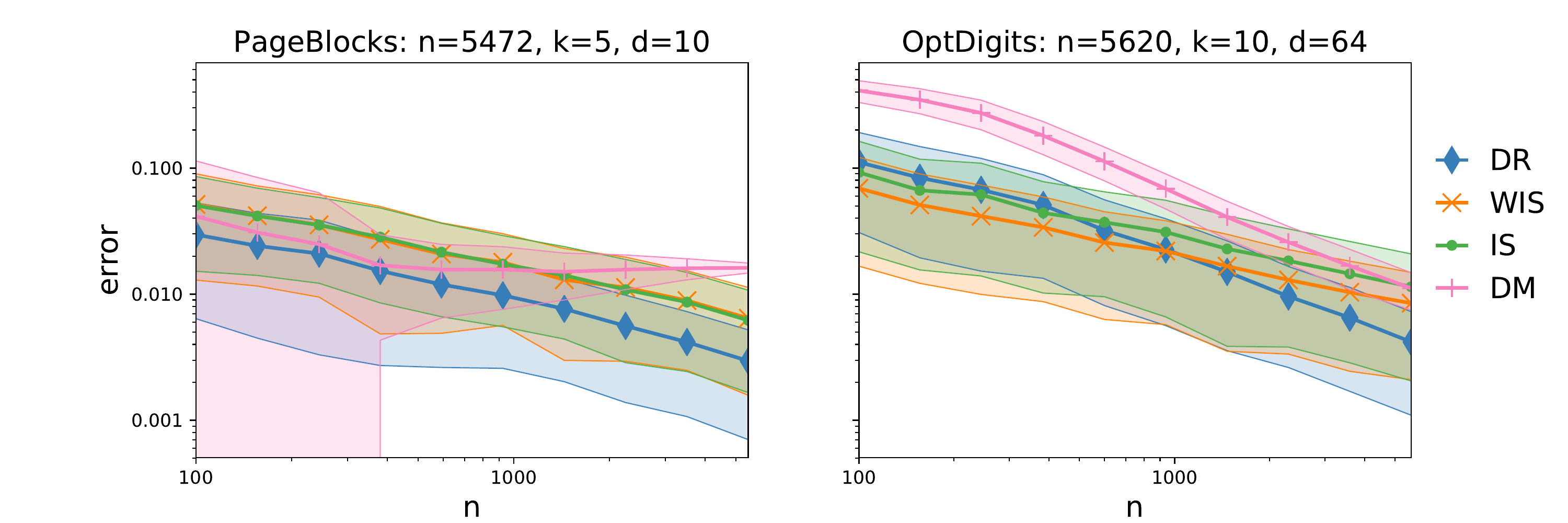}
    \caption{The error of the CDF estimators as a function of sample size $n$, for \textbf{(left)} the PageBlocks dataset and \textbf{(right)} the OptDigits dataset. Shaded area is the 95\% quantile over 500 runs. } 
    \label{fig:cdf_error}
    \vspace{-10px}
\end{figure}

\paragraph{Setup.} 
Following \cite{dudik2011doubly, dudik2014doubly, wang2017optimal}, 
we obtain our off-policy contextual bandit datasets 
by transforming classification datasets.
The contexts are the provided features, 
and the actions correspond to the possible class labels. 
To obtain the evaluation policy $\pi$,
we use the output probabilities 
of a trained logistic regression classifier. 
The behavior policy is defined as 
$\beta = \alpha \pi + (1-\alpha)\pi_{\text{UNIF}}$, 
where $\pi_{\text{UNIF}}$ 
is a uniform policy over the actions,
for some $\alpha \in (0, 1]$. 
We apply this process to the PageBlocks 
and OptDigits datasets~\cite{Dua:2019}, 
which have dimensions $d$ and actions $k$ 
using $\alpha = 0.1$
(Figure~\ref{fig:cdf_error}). 
When models $\overline{G}$ are used 
(for DM, DR estimators), 
as in \cite{dudik2011doubly},
the dataset is divided into two splits, 
with each of the two splits 
used to calculate $\overline{G}$ via regression, 
which is then used 
with the other split 
to calculate the estimator. 
The two results are averaged 
to produce the final estimators. 
We provide further details 
and extensive evaluations 
in Appendix~\ref{appendix:experiment}.

\paragraph{CDF Estimation.} 
We evaluate the error $\Vert F - \wh F\Vert_\infty$
of our \CDF estimators against sample size
for two UCI datasets (Figure \ref{fig:cdf_error}). 
The \IS and \DR exhibit the expected 
$O\left(1/\sqrt{n}\right)$ rate of convergence 
in error previously derived 
in Theorems~\ref{thm:is-cdf} 
and~\ref{thm:dr_error}, respectively. 
We note that the \WIS estimator, while biased,
performs as well as the \IS estimator if not better. 
In the PageBlocks dataset (Figure \ref{fig:cdf_error}, left),
the regression model for $\overline{G}$ 
is relatively well-specified 
as exemplified by the relatively low error of the $\DM$ estimator, though it has high variance for low samples sizes. 
The \DR estimator leverages this model to outperform 
all other estimators for all sample sizes, 
without suffering the drawbacks of the \DM estimator. 
It takes an order of magnitude less data
to reach the same error 
compared to the \IS and \WIS estimators. 
In contrast, the regression model 
is less well-specified in the OptDigits dataset for lower sample sizes
(Figure \ref{fig:cdf_error}, right), 
and consequently, 
the $\DR$ estimator cannot perform as well 
as the \IS and \WIS estimators for small $n$. 
This trend reverses as data increases and the model improves,
with the \DR estimator outperforming 
the \IS estimators.

\begin{figure}[t!]
\centering
\hspace*{-1cm}\includegraphics[width=\textwidth,]{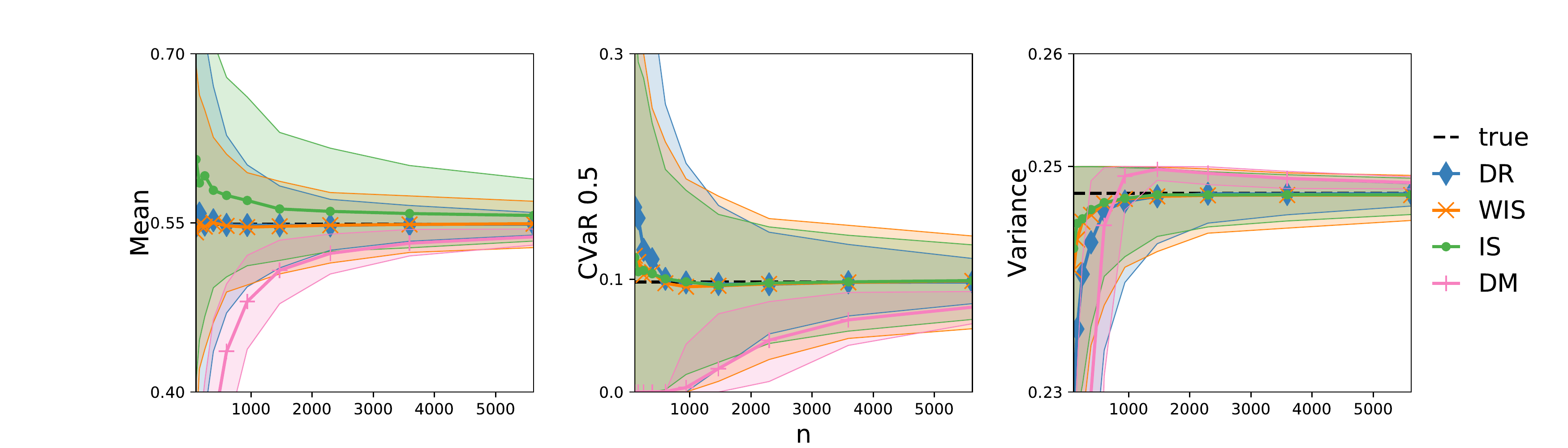}
\caption{Estimated mean, $\text{CVaR}_{0.5}$, and variance for the OptDigits dataset, compared to their true values (black). Shaded area is the standard deviation over 500 runs. }
\label{fig:risk_estimates}
\vspace{-10px}
\end{figure}

\paragraph{Estimation of Risk Functionals.}
Figure \ref{fig:risk_estimates} shows the 
mean, variance, and $\text{CVaR}_{0.5}$ estimates, 
which are obtained by evaluating each risk functional
on the \CDF estimators for the OptDigits dataset.
Here, the estimates are plotted against the true value (dashed line) 
to make the variance reduction effect of the \DR estimators more apparent. 
The \DM estimator, which appeared to have competitive performance in the \CDF error plot,
has relatively high risk estimate error, 
which occurs because the \DM \CDF may be poorly approximated 
in areas that are important for risk functional estimation. 
The \IS, \WIS, and \DR risk estimates 
converge quickly to the true value as $n$ increases,
and as expected, 
their relative behavior echoes the trends in Figure~\ref{fig:cdf_error} as a consequence of our distributional approach. 
The \DR estimator has slightly worse performance for small samples sizes due to the poor specification of the model, but soon exhibits 
the desired variance reduction for $n > 1000$.

\section{Discussion}\label{sec:conclusion} 
In this paper, we have developed a distribution-centric method for high confidence off-policy estimation of risk functionals. 
Our method relies on first estimating the CDF and its confidence band, then estimating risk functionals by evaluating on the estimated CDF. 
We have defined several estimators for the CDF, including an importance sampling and doubly robust estimator which takes advantage of side information to reduce variance.  
For $L$-Lipschitz risk functionals, which we show many classes of risks fall under, the concentration of the estimated risk can be derived from the confidence band on the CDF. 

From a theoretical point of view, our paper provides the first finite sample concentration inequalities for a number of different CDF and risk estimators, which are widely applicable to recent distributional reinforcement learning settings, which learn the CDF of returns and are capable of optimizing different risk functionals~\citep{dabney2018implicit, keramati2020being}. 
Of these estimators, the doubly robust estimator is a novel contribution and has not yet been defined or analyzed for distributions in the literature. 
From a practical standpoint, our method 
can be used to comprehensively evaluate
the behavior of a target policy before deployment 
using a wide range of risk functionals--a contribution that is especially important in real-world applications. 

Our work also raises 
several open questions and avenues of future work, 
which we discuss below. 

\paragraph{Error Bound for the Doubly Robust CDF Estimator.}
Although we presented a sample-dependent bound, we believe that it can be improved because our current bound does not take into account the double robustness of the estimator. 
We obtain a looser bound for the DR estimator compared to the IS estimator, even though we show the DR estimator has reduced pointwise variance. 
Obtaining an improved error bound dependent on the term $G-\wb{G}$, which is the error of the given model compared to the true conditional \CDF, is one important direction of future work. 

\paragraph{Monotone Transformation of the CDF Estimate.} 
As we have demonstrated for 
the importance sampling and doubly robust estimators, 
\CDF estimation faces a unique problem in that the 
estimate may not be a valid \CDF. 
Estimates of the expected value, 
for example, 
are not subject to any such constraints. 
We have shown how methods 
such as clipping and monotone transformation 
can be applied to $\wh{F}_{\text{IS}}$ or $\wh{F}_{\text{DR}}$ 
to mitigate this problem. 

However, it is important to note that 
there are, in fact, several options 
for how the clipping and monotone transformation is applied. 
For example, instead of applying these transformations after averaging the $n$ samples to form the estimator, 
another option is to apply the transformation to each individual sample, 
and then average the transformed results. 
Applying the monotone transformation to each sample before averaging may potentially increase bias while reducing variance, which may be desirable in certain applications. 

In this paper we proposed a simple method~\eqref{eq:dr_monotone} for clipping and transforming the \CDF{} estimate, but different forms of monotone regression \citep{smith1987best, darst1985best}, 
which may potentially provide a better monotone approximation of the CDF estimate. 
As of yet, the best method of transforming estimates into valid \CDF{}s is not yet clear. 
Extensive theoretical and empirical evaluation of such estimator options is another important avenue of future work. 

\paragraph{CDF and Risk Estimation in MDPs.} Previous work in off-policy evaluation for expected value has developed a doubly robust estimator for the MDP setting~\citep{jiang2016doubly}. Following this, another avenue of future work will aim to extend our results for the contextual bandit setting to CDF and risk estimation in the Markov Decision Process (MDP) and nonstationary settings. 
We believe this is especially relevant in relation to recent advances in distributional reinforcement learning, that aims to learn the distribution of returns in MDPs~\citep{dabney2018implicit}. 

\paragraph{CDF and Risk Estimation with Unknown Behavioral Policy.}
Along similar lines, another direction of future work lies in in-depth analysis of the importance sampling and doubly robust estimators when the behavioral policy is unknown and must be estimated from the data. Previous works have shown that under mild consistency assumptions, using estimates of importance weight asymptotically provides a better estimator of policy evaluate~\citep{hahn1998role,hirano2003efficient,heckman1998characterizing}. Whether similar properties hold under CDF and risk estimation remains to be seen.

\paragraph{Risk Error Bounds Without Lipschitzness.}Finally, though we provide concentration bounds for a large number of risk functionals under the $L$-Lipschitz property, 
a number of other risk functionals, such as the inverse quantile function, do not satisfy this property.
The confidence band on the CDF can still be used to calculate a confidence interval on the risk, 
but it is not clear if and how quickly the confidence interval shrinks with more samples. 
This motivates the following open question, which we plan to study in future work: can concentration inequalities for risk functional estimates be derived if and only if they are Lipschitz, in a more general sense?

\section*{Acknowledgements}
The authors thank Siva Balakrishnan for inspiring discussions, 
and Roberto Imbuzeiro Oliveira for his lecture notes. 
Liu Leqi is generously supported by an Open Philanthropy AI Fellowship. Zachary Lipton thanks Amazon AI, Salesforce Research, the Block Center, the PwC Center, Abridge, UPMC, the NSF, DARPA, and SEI for supporting ACMI lab’s research on robust and socially aligned machine learning.

\newpage
\bibliographystyle{plain}
\bibliography{refs.bib}

\newpage
\appendixwithtoc
\newpage

\appendix

\section{Proofs for Risk Functionals (Section~\ref{sec:risk_functionals})}\label{appendix:lipschitz_risk}

\begin{proof}[Proof of Lemma \ref{lem:distorted_lipschitz}]
\begin{align*}
    \vert \rho\left(F_Z\right) - \rho\left(F_{Z'}\right) \vert &= \left\vert \int_0^D g\left(1-F_Z(t)\right) - g\left(1-F_{Z'}(t)\right) dt \right\vert \\
    &\leq \int_0^D \left\vert g\left(1-F_Z(t)\right) - g\left(1-F_{Z'}(t)\right) \right\vert dt \\ 
    &\leq \int_0^D \frac{L}{D} \left\vert F_{Z'}(t) - F_Z(t) \right\vert dt \\ 
    &\leq L \max_{t} \left\vert F_{Z}(t) - F_{Z'}(t) \right\vert,  
\end{align*}
where the second to last step uses the $L/D$-Lipschitzness of $\rho$. 
\end{proof}

\begin{proof}[Proof of Lemma~\ref{lem:cpt_lipschitz}] 
Using the definition of the \CDF, note that on the bounded support of $[0, D]$ the CPT functional can be rewritten as 
\begin{align*}
    \rho(F_Z) = \int_0^{D}g^+\left(\Prob_Z\left(u^+(Z)  > t \right)\right) dt - \int_0^{D}g^-\left(\Prob_Z\left(u^-(Z)  > t \right)\right) dt. 
\end{align*}
Then, 
\begin{align*}
    \left\vert \rho(Z) - \rho(Z') \right\vert &= \Big\vert \int_0^{D}g^+\left(\Prob_Z\left(u^+(Z)  > t \right)\right) dt - \int_0^{D}g^-\left(\Prob_Z\left(u^-(Z)  > t \right)\right) dt \\
    &\quad\quad- \int_0^{D}g^+\left(\Prob_{Z'}\left(u^+(Z')  > t \right)\right) dt - \int_0^{D}g^-\left(\Prob_{Z'}\left(u^-(Z')  > t \right)\right) dt \Big\vert \\ 
    &\leq \left\vert \int_0^{D}g^+\left(\Prob_Z\left(u^+(Z)  > t \right)\right) dt - \int_0^{D}g^+\left(\Prob_{Z'}\left(u^+(Z')  > t \right)\right) dt \right\vert \\
    &\quad\quad+ \left\vert \int_0^{D}g^-\left(\Prob_{Z}\left(u^-(Z)  > t \right)\right) dt - \int_0^{D}g^-\left(\Prob_{Z'}\left(u^-(Z')  > t \right)\right) dt \right\vert \\ 
    &\leq \frac{L}{D}\int_0^{D}\left\vert \Prob_Z\left(u^+(Z)  > t \right) - \Prob_{Z'}\left(u^+(Z')  > t \right) \right\vert  dt \\
    &\quad\quad+ \frac{L}{D}\int_0^{D}\left\vert \Prob_Z\left(u^-(Z)  > t \right) - \Prob_{Z'}\left(u^-(Z')  > t \right) \right\vert dt \\
    &\leq \frac{L}{D}\int_0^{D}\left\vert \Prob_Z\left(Z  > t \right) - \Prob_{Z'}\left(Z'  > t \right) \right\vert  dt \\
    &\quad\quad+ \frac{L}{D}\int_0^{D}\left\vert \Prob_Z\left(Z  > t \right) - \Prob_{Z'}\left(Z'  > t \right) \right\vert dt \\
    &= 2\frac{L}{D}\int_0^{D}\left\vert F_{Z'}(t) - F_Z(t) \right\vert  dt \\
    &\leq 2L\max_t \left\vert F_{Z}(t) - F_{Z'}(t) \right\vert
\end{align*}
\end{proof}

\begin{proof}[Proof of Lemma \ref{lemma:variance_lipschitz}]
For the variance of any random variable $Z$ with bounded support $[0,D]$, we have
\begin{align*}
    \Var(Z) = \E(Z^2) - \E(Z)^2.
\end{align*}
Note that by the definition of expectation, 
\begin{align*}
    \E(Z^2) &= \int_{t^2 = 0}^{D^2} 1 - F_{Z^2}(t^2)d t^2
\end{align*}
Then using $dt^2 = 2tdt$ and the fact that $\Prob(Z^2 \geq t^2) = \Prob(Z \geq t)$ since $t$ is nonnegative, with this change of variables we have 
\begin{align*}
    \E(Z^2) &= 2\int_{t = 0}^{D} t \left(1 - F_Z(t)\right) dt.  
\end{align*}
This gives us the following expression for variance:
\begin{align*}
    \Var(Z) = 2\int_{0}^D t(1-F_Z(t))dt - \left(\int_0^D (1-F_Z(t))dt \right)^2
\end{align*}
Next, consider a pair of random variables $Z$ and $Z'$ with $F_Z$ and $F_{Z'}$ as their \CDF respectively. Therefore,  
\begin{align*}
    \left|\Var(Z) - \Var(Z') \right|
    & \leq \left| 2\int_{0}^D t(F_Z(t)-F_{Z'}(t))dt\right| +\left|  \left(\int_0^D (1-F_Z(t))dt \right)^2- \left(\int_0^D (1-F_{Z'}(t))dt \right)^2\right|\\
    & \leq D^2\|F_Z(t)-F_{Z'}\|_{\infty} +\left|  \int_0^D (F_Z(t)-F_{Z'}(t))dt \right| \left|\int_0^D (1-F_{Z}(t))dt+\int_0^D (1-F_{Z'}(t))dt \right|\\
    &\leq D^2\|F_Z(t)-F_{Z'}\|_{\infty} + 2D\left|  \int_0^D (F_Z(t)-F_{Z'}(t))dt \right| \\
    & \leq D^2\|F_Z(t)-F_{Z'}\|_{\infty} +2D^2\|F_Z(t)-F_{Z'}\|_{\infty} \\
    &=   3D^2\|F_Z-F_{Z'}\|_{\infty}
\end{align*}
\end{proof}

\begin{proof}[Proof of Lemma \ref{lem:weighted_lipschitz}]
The proof of this lemma follows directly from the definition of Lipschitzness: 
\begin{align*}
    \left\vert \sum_{k=1}^K \lambda_k \rho_k(Z) - \sum_{k=1}^K \lambda_k \rho_k(Z') \right\vert &\leq  \sum_{k=1}^K \lambda_k \left\vert  \rho_k(Z) - \rho_k(Z') \right \vert \\
    &\leq \Vert F_Z - F_{Z'} \Vert_{\infty} \sum_{k=1}^K \lambda_k L_k. 
\end{align*}

\end{proof}
\clearpage

\section{Proofs for CDF Estimation (Section~\ref{sec:cdf_estimator})}
\subsection{Importance Sampling (IS) Estimators (Section~\ref{sec:is_estimator})} 

\subsubsection{Proof: Bias and Variance of IS CDF Estimate}
\begin{proof}[Proof of Lemma~\ref{lem:is_bias_var}]
We take the expectation of the IS estimator~\eqref{is_estimator} with respect to $\Prob_\beta$. 
Then for any $t \in \mathbb{R}$, 
\begin{align*}
    \E_{\mathbb{P}_\beta}[\wh F_\IS(t)]
    &= \E_{\mathbb{P}_\beta}\left[\frac{1}{n}\sum_{i=1}^n
    w(A_i, X_i)\Ind_{\{R_i \leq t\}}\right]\\
    &= \E_{\Prob_\beta} \left[\E_{\Prob_\beta} \left[
    \frac{\pi(A|X)}{\beta(A|X)}
    \E_{\Prob_\beta}\left[
    \Ind_{\{R \leq t\}} | X, A\right]\right] \right]\\
    &= \E_{\mathbb{P}}\left[
    w(A, X)\Ind_{\{R \leq t\}}\right]\\
    &= F(t).
\end{align*}
Recall that $G(t; X,A) = \mathbb{E}[\Ind_{\{R \leq t\}} | X,A\}]$.
The variance of the IS estimator is derived using:
\begin{align*}
    \Var_{\Prob_\beta}\left[\wh{F}_{\text{IS}}(t)\right] &= \frac{1}{n} \Var_{\Prob_\beta}\left[w(A,X)\mathbbm{1}_{\lbrace R \leq t \rbrace} \right] \\ 
    &= \frac{1}{n}\E_{\Prob_\beta}\left[w(A,X)^2\Var_{\Prob_\beta}\left[\mathbbm{1}_{\lbrace R \leq t \rbrace} | A,X\right]\right] + \frac{1}{n}\Var_{\Prob_\beta}\left[w(A,X)\E_{\Prob_\beta}\left[ \mathbbm{1}_{\lbrace R \leq t \rbrace} | A,X\right] \right] \\ 
    &= \frac{1}{n}\E_{\Prob_\beta}\left[w(A,X)^2 \sigma^2(t;X,A) \right] + \frac{1}{n}\Var_{\Prob_\beta}\left[w(A,X) G(t;X,A)\right] \\ 
    &= \frac{1}{n}\E_{\Prob_\beta}\left[w(A,X)^2 \sigma^2(t;X,A) \right] + \frac{1}{n}\Var_{\Prob_\beta}\left[\E_{\Prob_\beta}\left[w(A,X) G(t;X,A)\vert X\right]\right] \\ 
    &\quad\quad+ \frac{1}{n}\E_{\Prob_\beta}\left[\Var_{\Prob_\beta}\left[w(A,X) G(t;X,A)\vert X\right]\right] 
\end{align*}
where the second equality uses the law of total variance conditioned on actions $A$ and contexts $X$, and the third equality uses the definitions of $\sigma^2$ and $G$.  
The last equality is another application of the law of total variance conditioning on the context $X$. 

\end{proof}

\subsubsection{Proof: Error Bound of IS CDF Estimate}\label{proof:is-error}
\begin{proof}[Proof Theorem \ref{thm:is-cdf}]

Define the following function class: 
\begin{align*}
    \FuncClass(n):=\Big\lbrace f(r) :=\varrho\frac{1}{n}\Ind_{\lbrace r\leq t\rbrace}: \forall t\in\Real;\forall r\in\Q, \varrho\in\lbrace -1,+1\rbrace\Big\rbrace
\end{align*}
Note that this is a countable set. Using this definition, we have
\begin{align*}
    \sup_{t\in\Real} \left|\wh F_\IS(t) -  F(t)\right| =\sup_{f\in\FuncClass(n)}\left|\left(\sum_i^n \left(\weight(A_i,X_i)f(R_i)-\E_{\Prob_\beta}[\weight(A_i,X_i)f(R_i)]\right)\right)\right|
\end{align*}
Using this equality, for $\lambda>0$, we have:
\allowdisplaybreaks
\begin{align*}
    \E_{\Prob_\beta}&\left[\exp\left(\lambda\sup_{t\in\Real} \left|\wh F_\IS(t) -  F(t)\right|\right) \right]\\
    &\quad\quad\quad=\E_{\Prob_\beta}\left[\exp\left(\lambda\sup_{f\in\FuncClass(n)}\left| \left(\sum_i^n\left( \weight(A_i,X_i)f(R_i)-\E_{\Prob_\beta}[\weight(A_i,X_i)f(R_i)]\right)\right)\right|\right) \right]\\
    &\quad\quad\quad=\E_{\Prob_\beta}\left[\exp\left(\lambda\sup_{f\in\FuncClass(n)}\left| \left(\E_{\Prob_\beta}\left[\sum_i^n \left(\weight(A_i,X_i)f(R_i)-\weight(X_i',A_i')f(R_i')\right)\Big|
    \lbrace X_i,A_i,R_i\rbrace_i^n
    \right]\right)\right|\right) \right]\\
    &\quad\quad\quad\leq\E_{\Prob_\beta}\left[\exp\left(\lambda\sup_{f\in\FuncClass(n)}\left| \E_{\Prob_\beta}\left[\left(\sum_i^n \left(\weight(A_i,X_i)f(R_i)-\weight(X_i',A_i')f(R_i')\right)\Big|
    \lbrace X_i,A_i,R_i\rbrace_i^n
    \right]\right)\right|\right) \right]\\
    &\quad\quad\quad\leq\E_{\Prob_\beta}\left[\exp\left(\lambda\E_{\Prob_\beta}\left[\sup_{f\in\FuncClass(n)}\left| \left(\sum_i^n \left(\weight(A_i,X_i)f(R_i)-\weight(X_i',A_i')f(R_i')\right)\Big|
    \lbrace X_i,A_i,R_i\rbrace_i^n
    \right]\right)\right|\right) \right]\\
    &\quad\quad\quad\leq\E_{\Prob_\beta}\left[\exp\left(\lambda\sup_{f\in\FuncClass(n)}\left| \left(\sum_i^n \left(\weight(A_i,X_i)f(R_i)-\weight(X_i',A_i')f(R_i')\right)\right)\right|\right) \right]\\
    &\quad\quad\quad=\E_{\Prob_\beta,\R}\left[\exp\left(\lambda\sup_{f\in\FuncClass(n)}\left| \left(\sum_i^n \xi_i(\weight(A_i,X_i)f(R_i)-\weight(X_i',A_i')f(R_i'))\right)\right|\right) \right]\\
    &\quad\quad\quad\leq\E_{\Prob_\beta,\R}\left[\exp\left(2\lambda\sup_{f\in\FuncClass(n)}\left|\left(\sum_i^n \xi_i\weight(A_i,X_i)f(R_i)\right)\right|\right) \right]\\
    &\quad\quad\quad=\E_{\Prob_\beta,\R}\left[\sup_{f\in\FuncClass(n)}\exp\left(2\lambda\left|\left(\sum_i^n \xi_i\weight(A_i,X_i)f(R_i)\right)\right|\right) \right]
\end{align*}
with $\R$ a Rademacher measure on a set of Rademacher random variable $\{\xi_i\}$ a Rademacher random variable.

Next, permute the indices $i$ such that $R_1 \leq \ldots R_i \ldots \leq R_n$.
Consider a function $f(r) = \frac{1}{n}\varrho\Ind_{\lbrace r\leq t\rbrace}$. 
For such a function, $\sum_i^n \xi_i \weight(A_i,X_i)f(R_i)$ is equal to 
\begin{itemize}
    \item $0$ if $t<\min_i\lbrace R_i\rbrace_i^n$,
    \item $\frac{1}{n}\varrho\sum_i^j\weight(A_i,X_i)\xi_i$ when $R_j\leq t<R_{j+1}$ for a $j\in\lbrace1,\ldots,n-1\rbrace$,
    \item $\frac{1}{n}\varrho\sum_i^n\weight(A_i,X_i)\xi_i$ otherwise. 
\end{itemize}
Then,
\begin{align*}
    \sup_{f\in\FuncClass(n)}\exp&\left(2\lambda\left|\left(\sum_i^n \xi_i\weight(A_i,X_i)f(R_i)\right)\right|\right) \\
    &\quad\quad\quad= \max_{\varrho,j} \exp\left(\frac{2\lambda}{n}\varrho\sum_i^j\weight(A_i,X_i)\xi_i\right)\\
    &\quad\quad\quad= \max_{j}\Big( \exp\left(\frac{2\lambda}{n}\sum_i^j\weight(A_i,X_i)\xi_i\right)\Ind_{\lbrace\sum_i^j\weight(A_i,X_i)\xi_i\geq0\rbrace}\\
    &\quad\quad\quad\quad\quad\quad\quad\quad\quad+\exp\left(-\frac{2\lambda}{n}\sum_i^j\weight(A_i,X_i)\xi_i\right)\Ind_{\lbrace\sum_i^j\weight(A_i,X_i)\xi_i<0\rbrace}\Big)\\
    &\quad\quad\quad= \max_{j}\left( \exp\left(\frac{2\lambda}{n}\sum_i^j\weight(A_i,X_i)\xi_i\right)\Ind_{\lbrace\sum_i^j\weight(A_i,X_i)\xi_i\geq0\rbrace}\right)\\
    &\quad\quad\quad\quad\quad\quad\quad\quad\quad+\max_{j}\left(\exp\left(-\frac{2\lambda}{n}\sum_i^j\weight(A_i,X_i)\xi_i\right)\Ind_{\lbrace\sum_i^j\weight(A_i,X_i)\xi_i<0\rbrace}\right)
\end{align*}

Which gives us the inequality

\begin{align}\label{eq:symmetrization}
    \E_{\Prob_\beta}\left[\exp\left(\lambda\sup_{t\in\Real} \left|\wh F_\IS(t) -  F(t)\right|\right) \right]&\leq
    2\E_{\Prob_\beta,\R}\left[ \max_{j}\exp\left(\frac{2\lambda}{n}\sum_i^j\weight(A_i,X_i)\xi_i\right) \Ind_{\lbrace\sum_i^j\weight(A_i,X_i)\xi_i\geq 0\rbrace}\right]
\end{align}
Now we are left to bound the right hand side of \eqref{eq:symmetrization}. 
Using 
Lemma~\ref{lemma:CDF_Convex}, for the right hand side of the \eqref{eq:symmetrization} we have,

\begin{align}\label{eq:cdf_trick}
    &\E_{\Prob_\beta,\R}\left[ \exp\left(\frac{2\lambda}{n}\max_{j}\sum_i^j\weight(A_i,X_i)\xi_i\right) \Ind_{\lbrace\max_{j}\sum_i^j\weight(A_i,X_i)\xi_i\geq 0\rbrace}\right]\nonumber\\
    &\quad\quad\quad\quad=\Prob_\beta\lbrace \max_{j}\frac{2\lambda}{n}\sum_i^j\weight(A_i,X_i)\xi_i\geq 0\rbrace\nonumber\\
    &\quad\quad\quad\quad\quad\quad\quad\quad+\lambda\int_0^\infty\exp(\lambda t)\Prob\lbrace \max_{j}\frac{2\lambda}{n}\sum_i^j\weight(A_i,X_i)\xi_i\geq t\rbrace dt\nonumber\\
    &\quad\quad\quad\quad\leq\Prob_\beta\lbrace \max_{j}\frac{2\lambda}{n}\sum_i^j\weight(A_i,X_i)\xi_i\geq 0\rbrace\nonumber\\
    &\quad\quad\quad\quad\quad\quad\quad\quad+2\lambda\int_0^\infty\exp(\lambda t)\Prob\lbrace \frac{2\lambda}{n}\sum_i\weight(A_i,X_i)\xi_i\geq t\rbrace dt
\end{align}
Note that similarly we have,

\begin{align}\label{eq:trick}
    &\E_{\Prob_\beta,\R}\left[ \exp\left(\frac{2\lambda}{n}\sum_i\weight(A_i,X_i)\xi_i\right) \Ind_{\lbrace\sum_i\weight(A_i,X_i)\xi_i\geq 0\rbrace}\right] \nonumber\\
    =&\Prob_\beta\lbrace \frac{2\lambda}{n}\sum_i\weight(A_i,X_i)\xi_i\geq 0\rbrace
    +\lambda\int_0^\infty\exp(\lambda t)\Prob\lbrace \frac{2\lambda}{n}\sum_i\weight(A_i,X_i)\xi_i\geq t\rbrace dt
\end{align}
Putting these two statements, i.e.,~\eqref{eq:cdf_trick}, and \eqref{eq:trick} together, and applying the result of Lemma~\ref{lemma:maxProb}, we have,

\begin{align*}
    &\E_{\Prob_\beta,\R}\left[ \exp\left(\max_{j}\frac{2\lambda}{n}\sum_i^j\weight(A_i,X_i)\xi_i\right) \Ind_{\lbrace\sum_i^j\weight(A_i,X_i)\xi_i\geq 0\rbrace}\right]\\
    &\quad\quad\quad\quad
    \leq\Prob_\beta\lbrace \max_{j}\frac{2\lambda}{n}\sum_i^j\weight(A_i,X_i)\xi_i\geq 0\rbrace\\
    &\quad\quad\quad\quad\quad\quad\quad\quad+2\E_{\Prob_\beta,\R}\left[ \exp\left(\frac{2\lambda}{n}\sum_i\weight(A_i,X_i)\xi_i\right) \Ind_{\lbrace\sum_i\weight(A_i,X_i)\xi_i\geq 0\rbrace}\right]\\
    &\quad\quad\quad\quad\quad\quad\quad\quad\quad\quad\quad\quad-2\Prob_\beta\lbrace \frac{2\lambda}{n}\sum_i\weight(A_i,X_i)\xi_i\geq 0\rbrace\\
    &\quad\quad\quad\quad\leq 2\E_{\Prob_\beta,\R}\left[ \exp\left(\frac{2\lambda}{n}\sum_i\weight(A_i,X_i)\xi_i\right) \Ind_{\lbrace\sum_i\weight(A_i,X_i)\xi_i\geq 0\rbrace}\right]\\
    &\quad\quad\quad\quad\leq 2\E_{\Prob_\beta,\R}\left[ \exp\left(\frac{2\lambda}{n}\sum_i\weight(A_i,X_i)\xi_i\right)\right]\\
\end{align*}

Note that $\frac{2}{n}\weight(A_i,X_i)\xi_i$ is a mean zero random variable with values in $[-\frac{2 }{n}\weight_{\max},\frac{2 }{n}\weight_{\max}]$. Therefore, it is a sub-Gaussian random variable with sub-Gaussian constant as $\left(\frac{2 }{n}\right)^2\weight_{\max}^2$. Using this, we have, $\frac{2}{n}\sum_i\weight(A_i,X_i)\xi_i$ is $\frac{4 }{n}\weight_{\max}^2$ sub-Gaussian random variable. Therefore, we have,

\begin{align*}
    \E_{\Prob_\beta,\R}\left[ \exp\left(\max_{j}\frac{2\lambda}{n}\sum_i^j\weight(A_i,X_i)\xi_i\right) \Ind_{\lbrace\sum_i^j\weight(A_i,X_i)\xi_i\geq 0\rbrace}\right]
    &\leq 2\E_{\Prob_\beta,\R}\left[ \exp\left(\frac{2\lambda}{n}\sum_i\weight(A_i,X_i)\xi_i\right)\right]\\
    &\leq2 \exp\left(\lambda^2\frac{2}{n}\weight_{\max}^2\right)
\end{align*}

Putting this with the \eqref{eq:symmetrization}, we have 

\begin{align*}
    \E_{\Prob_\beta}\left[\exp\left(\lambda\sup_{t\in\Real} \left|\wh F_\IS(t) -  F(t)\right|\right) \right]&\leq
    2\E_{\Prob_\beta,\R}\left[ \exp\left(\max_{j}\frac{2\lambda}{n}\sum_i^j\weight(A_i,X_i)\xi_i\right) \Ind_{\lbrace\sum_i^j\weight(A_i,X_i)\xi_i\geq 0\rbrace}\right]\\
    &\leq 4 \exp\left(\lambda^2\frac{2}{n}\weight_{\max}^2\right)
\end{align*}
Using Markov inequality we have

\begin{align*}
    \Prob_\beta\left(\sup_{t\in\Real} \left|\wh F_\IS(t) -  F(t)\right|\geq \epsilon\right)&=\Prob_\beta\left(\exp\left(\lambda\sup_{t\in\Real} \left|\wh F_\IS(t) -  F(t)\right|\right)\geq \exp(\lambda\epsilon)\right)\\
    &\leq 4 \exp\left(\lambda^2\frac{2}{n}\weight_{\max}^2\right)\exp(-\lambda\epsilon)\\
    &=4 \exp\left(\lambda^2\frac{2}{n}\weight_{\max}^2-\lambda\epsilon\right)
\end{align*}

This holds for any choice of $\lambda>0$, resulting in 
\begin{align*}
    \Prob_\beta\left(\sup_{t\in\Real} \left|\wh F_\IS(t) -  F(t)\right|\geq \epsilon\right)&\leq\inf_{\lambda>0}4 \exp\left(\lambda^2\frac{2}{n}\weight_{\max}^2-\lambda\epsilon\right)=4 \exp\left(\frac{-n\epsilon^2}{8\weight_{\max}^2}\right)
\end{align*}

Using this, we have 

\begin{align*}
    \Prob_\beta\left(\sup_{t\in\Real} \left|\wh F_\IS(t) -  F(t)\right|\leq \sqrt{\frac{8\weight^2_{\max}}{n}\log\left(\frac{4}{\delta}\right)}\right) \geq 1-\delta.
\end{align*}

\textbf{Bernstein style:}
To bound this $\E_{\Prob_\beta,\R}\left[ \exp\left(\frac{2\lambda}{n}\sum_i\weight(A_i,X_i)\xi_i\right)\right]$ now we use Bernstein's. 
As discussed, the random variable $\weight(A_i,X_i)\xi_i$ is in $[-\weight_{\max},\weight_{\max}]$. However, if we look at its variance, we have $\E_{\Prob_\beta,\R}\left[\weight(A_i,X_i)^2\xi_i^2\right]=\E_{\Prob_\beta,\R}\left[\weight(A_i,X_i)^2\right]$ which is the second order R\'enyi divergence $d(\Prob||\Prob_\beta)$. Therefore, for $0<\lambda<\frac{n}{2\weight_{\max}}$, we have

\begin{align*}
    \E_{\Prob_\beta,\R}\left[ \exp\left(\frac{2\lambda}{n}\sum_i\weight(A_i,X_i)\xi_i\right)\right]&=\prod_i\E_{\Prob_\beta,\R}\left[ \exp\left(\frac{2\lambda}{n}\weight(A_i,X_i)\xi_i\right)\right]\\
    &\leq \prod_i \exp\left(\frac{\lambda^2\frac{4d(\Prob||\Prob_\beta)}{n^2}}{2\left(1-\lambda\frac{2}{n}\weight_{\max}\right)}\right)\\
    &=  \exp\left(\frac{n\lambda^2\frac{4d(\Prob||\Prob_\beta)}{n^2}}{2\left(1-\lambda\frac{2}{n}\weight_{\max}\right)}\right)
\end{align*}

Using the Markov inequality, we have,

\begin{align*}
    \Prob_\beta\left(\sup_{t\in\Real} \left|\wh F_\IS(t) -  F(t)\right|\geq \epsilon\right)
    &=4 \exp\left(\frac{n\lambda^2\frac{4d(\Prob||\Prob_\beta)}{n^2}}{2\left(1-\lambda\frac{2}{n}\weight_{\max}\right)}-\lambda\epsilon\right)
\end{align*}
Setting $\lambda = \frac{\epsilon}{\frac{2\weight_{\max}\epsilon}{n}+n\frac{4d(\Prob||\Prob_\beta)}{n^2}}$, we have,

\begin{align*}
    \Prob_\beta\left(\sup_{t\in\Real} \left|\wh F_\IS(t) -  F(t)\right|\geq \epsilon\right)
    &\leq 4 \exp\left(\frac{-\epsilon^2}{2\left(\frac{2}{n}\weight_{\max}\epsilon+n\frac{4d(\Prob||\Prob_\beta)}{n^2}\right)}\right)\\
    &= 4 \exp\left(\frac{-n\epsilon^2}{4\weight_{\max}\epsilon+8d(\Prob||\Prob_\beta)}\right)
\end{align*}

which results in,

\begin{align*}
    \Prob_\beta\left(\sup_{t\in\Real} \left|\wh F_\IS(t) -  F(t)\right|\leq \frac{4\weight_{\max}\log(\frac{4}{\delta})}{n}+2\sqrt{\frac{2d(\Prob||\Prob_\beta)\log(\frac{4}{\delta})}{n}}\right) \geq 1-\delta.
\end{align*}
Finally, we note that 
since $\sup_t \vert \widehat{F}_\text{IS-clip}(t) - F(t) \vert \leq \sup_t \vert \widehat{F}_\IS(t) - F(t) \vert$,
the above results for $\wh F_\IS$ also hold for $\wh F_\text{IS-clip}$.
\end{proof}

\subsection*{Auxiliary Lemmas}
\begin{lemma}
For any random variable $X$, with probability measure $\Prob$, we have
\begin{align*}
    \E\left[\exp(\lambda X)\Ind_{\{X\geq 0\}}\right] &=\Prob\lbrace X\geq 0\rbrace+\lambda\int_0^\infty\exp(\lambda t)\Prob\lbrace X\geq t\rbrace dt.
\end{align*}
\label{lemma:CDF_Convex}
\end{lemma}

\begin{proof}
for any random variable $X$, with probability measure $\Prob$, we have
\begin{align}\label{eq:CDF_Convex}
    \E\left[\exp(\lambda X)\Ind_{\{X\geq 0\}}\right] &= \E\left[\left(\exp(0)+\int_0^X\lambda\exp(\lambda t)dt\right)\Ind_{\{X\geq 0\}}\right]\nonumber\\
    &=\E\left[\Ind_{\{X\geq 0\}}\exp(0)\right]+\E\left[\Ind_{\{X\geq 0\}}\lambda\int_0^X\exp(\lambda t)\Ind_{\{X\geq 0\}}dt\right]\nonumber\\
    &=\Prob\lbrace X\geq 0\rbrace+\E\left[\lambda\int_0^X\exp(\lambda t)\Ind_{\{X\geq 0\}}dt\right]\nonumber\\
    &=\Prob\lbrace X\geq 0\rbrace+\lambda\int_0^\infty\exp(\lambda t)\Prob\lbrace X\geq t\rbrace dt.
\end{align}
\end{proof}

\begin{lemma}\label{lemma:maxProb}
For $\gamma>0$, we have,
\begin{align}\label{eq:maxProb}
    \Prob_\beta \left[\max_{j}\sum_i^j\weight(A_i,X_i)\xi_i\geq \gamma\right]\leq 2\Prob_\beta\left[\sum_i^n\weight(A_i,X_i)\xi_i\geq \gamma\right]
\end{align}
\end{lemma}

\begin{proof}\label{proof:maxProb}

Consider events $E_j:=\lbrace\sum_i^j\weight(A_i,X_i)\xi_i\geq \gamma,\sum_i^l\weight(A_i,X_i)\xi_i< \gamma, \forall l< j\rbrace$ with $E_0:=\emptyset$. Using these definitions, we have,
\begin{align*}
    \lbrace\max_{j}\sum_i^j\weight(A_i,X_i)\xi_i\geq \gamma\rbrace\subset\bigcup_j E_j
\end{align*}
Also,
\begin{align*}
 \bigcup_j \left(E_j\bigcap \lbrace\sum_{i>j}\weight(A_i,X_i)\xi_i\geq 0\rbrace\right) \subset \lbrace\sum_i\weight(A_i,X_i)\xi_i\geq \gamma\rbrace
\end{align*}

Also note that 
\begin{align*}
    \Prob_\beta\left[\sum_{i>j}\weight(A_i,X_i)\xi_i\geq 0\right]\geq \frac{1}{2}
\end{align*}
since this quantity is mean zero and symmetric.
Also note that the event 
 $\sum_{i>j}\weight(A_i,X_i)\xi_i$ is independent of $E_j$. 
 
 Using these, we have, 
\begin{align*}
\Prob_\beta\left[E_j\bigcap \lbrace\sum_{i>j}\weight(A_i,X_i)\xi_i\geq 0\rbrace\right] =\Prob_\beta\left[E_j\right]\Prob_\beta\left[ \lbrace\sum_{i>j}\weight(A_i,X_i)\xi_i\geq 0\rbrace\right]\geq\frac{\Prob_\beta\left[E_j\right]}{2} 
\end{align*}

As a result we have,
\begin{align*}
    \Prob_\beta\left[\sum_i\weight(A_i,X_i)\xi_i\geq \gamma\right]&\geq \Prob_\beta\left[\bigcup_j \left(E_j\bigcap \lbrace\sum_{i>j}\weight(A_i,X_i)\xi_i\geq 0\rbrace\right) \right]\\
    &=\sum_j\Prob_\beta\left[ E_j\bigcap \lbrace\sum_{i>j}\weight(A_i,X_i)\xi_i\geq 0\rbrace \right]\\
    &\geq \sum_j\frac{\Prob_\beta\left[E_j\right]}{2} \\
    &\geq \frac{\Prob_\beta\left[\lbrace\max_{j}\sum_i^j\weight(A_i,X_i)\xi_i\geq \gamma\rbrace\right]}{2}
\end{align*}
which concludes the statement.
\end{proof}

\clearpage

\subsection{Doubly Robust (DR) Estimators (Section~\ref{sec:dr_estimator})}

\subsubsection{Proof: Bias and Variance of DR CDF Estimate}\label{proof:dr_bias_variance}

\begin{proof}[Proof of Lemma~\ref{lem:dr_bias_var}]
The expectation of the DR estimator~\eqref{eq:dr_policy_est} is as follows:
\begin{align*}
  \E_{\Prob_\beta}\Big[\widehat F_\DR (t)\Big] &= \E_{\Prob_\beta}\Big[w(A, X) \mathbbm{1}_{\{R \leq t\}}\Big] 
  + \E_{\Prob_\beta}\Big[
  \overline{G}(t; X, \pi)
  - w(A, X) \overline{G}(t; X, A)
  \Big] \\
  &= F(t) + \E_{\Prob_\beta}\left[ \overline{G}(t;X,\pi) - \mathbb{E}_{\Prob_\beta}[w(A,X)\overline{G}(t;X,A) | X]\right]\\
  &= F(t) + \E_{\Prob_\beta}\left[ \overline{G}(t;X,\pi) - \overline{G}(t; X, \pi)\right]\\
  &= F(t).
\end{align*}
Next, we derive the variance. 
\begin{align*}
    \Var_{\Prob_\beta}\left[ \wh{F}_{\text{DR}}(t) \right] &= \frac{1}{n}\Var_{\Prob_\beta}\left[ w(A,X)\left( \mathbbm{1}_{\lbrace R \leq t \rbrace} - \overline{G}(t;X,A)\right) + \overline{G}(t;X,\pi)  \right] \\ 
    &= \frac{1}{n}\E_{\Prob_\beta}\left[ w(A,X)^2\sigma^2(t;X,A) \right] \\
    &\quad\quad+ \frac{1}{n}\Var_{ \Prob_\beta}\left[w(A,X)\left(G(t;X,A) - \overline{G}(t;X,A)\right) + \overline{G}(t;X,\pi)  \right] \\ 
    &= \frac{1}{n}\E_{\Prob_\beta}\left[ w(A,X)^2\sigma^2(t;X,A) \right] + \frac{1}{n}\Var_{\Prob_\beta}\left[\E_{\Prob_\beta}\left[ w(A,X)G(t;X,A) \vert X\right] \right] \\
    &\quad\quad+ \frac{1}{n}\E_{\Prob_\beta}\left[\Var_{\Prob_\beta}\left[ w(A,X)\left(G(t;X,A) - \overline{G}(t;X,A)\right) \vert X \right]\right]
\end{align*}
The first equality follows from applying the law of total variance, noting that the variance $\Var_{\Prob_\beta}\left[\overline{G}(t;X,A)|X,A\right] = 0$, and using the definitions of $G$ and $\sigma^2$.
The second equality again applies the law of total variance.
\end{proof}

\subsubsection{Proof: Error Bound of DR CDF Estimate}
\label{appendix:dr_error} 
\begin{proof}[Proof of Theorem~\ref{thm:dr_error}]
Recall that the DR estimator $\wh{F}_{\text{DR}}(t)$ is defined as 
\[
    \widehat{F}_\text{DR}(t) = \frac{1}{n}\sum_{i=1}^n w(a_i, x_i) \left( \mathbbm{1}_{\lbrace r_i \leq t \rbrace} - \overline{G}(t;x_i, a_i) \right) + \overline{G}(t;x_i,\pi)
\]
where $\overline{G}(t; x, \pi) = \sum_{a} \pi(a|x)\overline{G}(t; x,a)$. 
We can decompose the error of the DR estimator as:
\begin{align*}
    &\E_{\Prob_\beta}\left[\sup_t \vert \wh{F}_{\text{DR}}(t) - F(t)\vert \right] \\
    &\quad\quad= \E_{\Prob_\beta}\left[\sup_t \left\vert \left(\frac{1}{n}\sum_{i=1}^n w(A_i, X_i)\left(\mathbbm{1}_{\{R_i \leq t\}} - \overline{G}(t; X_i, A_i)\right) + \overline{G}(t; X_i, \pi) \right) - F(t) \right\vert\right] \\
    &\quad\quad\leq \E_{\Prob_\beta}\left[\sup_t \left( \left\vert \frac{1}{n}\sum_{i=1}^n w(A_i, X_i)\mathbbm{1}_{\{R_i \leq t\}} - F(t) \right\vert + \left\vert \frac{1}{n}\sum_{i=1}^{n}\overline{G}(t; X_i, \pi) - w(A_i, X_i)\overline{G}(t; X_i, A_i)\right\vert\right) \right]\\ 
    &\quad\quad\leq \E_{\Prob_\beta}\left[\sup_t \left\vert \frac{1}{n}\sum_{i=1}^n w(A_i, X_i)\mathbbm{1}_{\{R_i \leq t\}} - F(t) \right\vert + \sup_t \left\vert \frac{1}{n}\sum_{i=1}^{n}\overline{G}(t; X_i, \pi) - w(A_i, X_i)\overline{G}(t; X_i, A_i)\right\vert\right].
\end{align*}
We have already bounded the first term in Theorem \ref{thm:is-cdf},
and Lemma~\ref{lem:g_concentration} bounds the second term. Then in total, we have 
\begin{align*}
    \Prob_\beta \left( \sup_t \left\vert \wh{F}_\text{DR}(t) - F(t) \right\vert \geq \sqrt{\frac{8w_{max}^2}{n}\log\left(\frac{4}{\delta}\right)} + \sqrt{\frac{32w_{max}^2}{n}\log\frac{(2n)^{1/2}}{ w_{max}\delta}} \right) \leq 2\delta 
\end{align*}
Simplifying, 
\begin{equation}\label{eq:dr_error_2_result}
    \Prob_\beta \left( \sup_t \left\vert \wh{F}_{DR}(t) - F(t) \right\vert \geq \sqrt{\frac{72w_{max}^2}{n}\log\left(\frac{4n^{1/2}}{\delta}\right)}  \right) \leq 2\delta 
\end{equation}
which gives us our error bound for the DR estimator $\wh{F}_\text{DR}$.

As mentioned previously, however, $\wh{F}_{\text{DR}}$ may not be monotone, and in practice we must use a monotone transformation of the estimator. 
Consider a monotone transformation $\M$ of $\wh{F}_\text{DR}$ that is a simple accumulation function, e.g. $\forall t$, 
$$\M\left(\wh{F}_{DR}(t)\right) = \max_{t' \leq t}\wh{F}_{DR}(t')$$
Now we want to bound the error between the monotonized estimate $\M\left(\wh{F}_{DR}(t)\right)$ and $F$. Using our error bound in~\eqref{eq:dr_error_2_result}, let $\epsilon = \sqrt{\frac{72w_{max}^2}{n}\log\left(\frac{8n^{1/2}}{\delta}\right)}$. Then with probability at least $1-\delta$, for all $t \in \Real$, 
$$\max_t \vert \wh{F}_\text{DR}(t) - F(t) \vert \leq \epsilon.$$   
On this event, $\forall t$ there exists some $t' \leq t$ for which
\[
    \max_{t' \leq t}\widehat{F}_{DR}(t') - F(t) = \widehat{F}_{DR}(t') - F(t)  
\]
Using the fact that $F$ is monotone thus $F(t') \leq F(t)$, when $\widehat{F}_{DR}(t') \geq F(t)$ we have 
\[
\widehat{F}_{DR}(t') - F(t) \leq \widehat{F}_{DR}(t') - F(t') \leq \epsilon
\]
Similarly, when $\widehat{F}_{DR}(t') \leq F(t)$, 
\[
F(t) - \widehat{F}_{DR}(t') \leq F(t) - \widehat{F}_{DR}(t) \leq \epsilon
\]
Putting these two inequalities together, we have 
\[
\max_t \left\vert \mathcal{M}\left(\widehat{F}_{DR}\right)(t) - F(t) \right\vert \leq \epsilon. 
\]

The theorem statement, which applies to the clipped monotone transformation, follows from the fact that 
\[\max_t \left\vert \min\left\lbrace \mathcal{M}\left(\widehat{F}_{DR}\right)(t), 1 \right\rbrace - F(t) \right\vert \leq \max_t  \left\vert \M\left(\wh{F}_{DR}\right)(t) - F(t) \right\vert. 
\]

\end{proof}

\begin{lemma}\label{lem:g_concentration}
Let $\overline{G}(t;x,a)$ be a valid conditional \CDF for all $x \in \X, a \in \A$, and let $\weight: \mathcal{A} \times \mathcal{X} \rightarrow \Real$ be the importance sampling weights. Then for $\delta \in (0, 1]$, 
\begin{align*}
    \Prob_\beta\left(  \sup_t \left\vert \frac{1}{n}\sum_{i=1}^{n}\overline{G}(t; X_i, \pi) - \frac{1}{n}\sum_{i=1}^{n}w(A_i, X_i)\overline{G}(t; X_i, A_i) \right\vert \geq \sqrt{\frac{32w_{max}^2}{n}\log\frac{(2n)^{1/2}}{ w_{max}\delta}} \right) &\leq \delta.
\end{align*}
where $\overline{G}(t;x,\pi) = \E_{\Prob}[\overline{G}(t;x,A)|x]$. 
\end{lemma}

\begin{proof}
Since $\overline{G}$ is a valid \CDF,  
we apply Lemma \ref{Lemma:approximation} to $\overline{G}$.  
Consider a function of the form 
\[
 \overline\zeta(t;s^1, ..., s^m) = \frac{1}{m}\sum_{j=1}^m \mathbbm{1}_{\lbrace s^i \leq t \rbrace} 
\]
The function $\overline\zeta$ can be seen as a stepwise CDF function, where each step is $1/m$ and occurs at points $\lbrace s^j \rbrace_{j=1}^m$. 

Lemma~\ref{Lemma:approximation} approximates $\overline{G}$ using such $1/m$-stepwise \CDF{}s. 
For each context $x$ and action $a$, 
let $s^1_{x,a}, ..., s^m_{x,a} \in \mathbb{Q}^m$ be the points chosen according to the deterministic procedure in Lemma \ref{Lemma:approximation}, 
such that the following inequality holds:

\begin{align}
    \sup_t \left\vert \overline{G}(t; x, a) - \overline\zeta\left(t;\lbrace s^j_{x,a}\rbrace_{j=1}^m\right) \right\vert \leq \frac{1}{2m}. 
\end{align}\label{zetabar_ineq}

Next, consider the class of functions \begin{align*}
    \G(m) :=\Big\lbrace \zeta(s^1,...,s^m):=\frac{1}{m}\varrho\sum_{j=1}^{m}\Ind_{\lbrace s^j \leq t \rbrace}: \forall t\in\Real, \varrho \in \lbrace -1, +1 \rbrace ; \lbrace s^j\rbrace_{j=1}^m\in\Q^m\Big\rbrace
\end{align*}
Note that, $\overline\zeta$ is a subset of the function class $\G(m)$, e.g. $\overline\zeta \left(t; \lbrace s^j_{x,a}\rbrace_{j=1}^m\right) \in \G(m)$.

\begin{align*}
    &\sup_t \left\vert \frac{1}{n} \sum_{i=1}^n w(A_i, X_i)\overline{G}(t; X_i, A_i) -  \frac{1}{n}\sum_{i=1}^{n}\overline{G}(t; X_i, \pi) \right\vert\\ 
    &\quad\quad= \sup_{t} \left\vert \frac{1}{n} \sum_{i=1}^n w(A_i, X_i)\overline{G}(t; X_i, A_i) -  \frac{1}{n}\sum_{i=1}^{n}\E_{\Prob}\left[\overline{G}(t; X_i, A) \vert X_i \right]\right\vert\\
    &\quad\quad= \sup_{t} \left\vert \frac{1}{n} \sum_{i=1}^n w(A_i, X_i)\overline{G}(t; X_i, A_i) -  \frac{1}{n}\sum_{i=1}^{n}\E_{\Prob_\beta}\left[w(X_i,A)\overline{G}(t; X_i, A) \vert X_i \right]\right\vert\\
    &\quad\quad\leq \sup_{t} \left\vert \frac{1}{n}\sum_{i=1}^n w(A_i, X_i)\overline\zeta\left(t;\lbrace s^j_{X_i,A_i}\rbrace_{j=1}^m\right) - \frac{1}{n}\sum_{i=1}^n \E_{\Prob_\beta}\left[w(A, X_i)\overline\zeta\left(t;\lbrace s^j_{X_i,A}\rbrace_{j=1}^m\right) \Big\vert X_i\right] \right\vert + \frac{1}{m}\\
    &\quad\quad\leq \sup_{\zeta \in \mathcal{G}(m)} \left\vert \frac{1}{n}\sum_{i=1}^n w(A_i, X_i)\zeta(\lbrace s^j_{X_i,A_i}\rbrace_{j=1}^m) - \frac{1}{n}\sum_{i=1}^n \E_{\Prob_\beta}\left[w(A, X_i)\zeta(\lbrace s^j_{X_i,A}\rbrace_{j=1}^m) \Big\vert X_i\right] \right\vert + \frac{1}{m}
\end{align*}
where the second line uses the definition of $\overline{G}(t;X_i,\pi)$, 
the third line uses a change of measure through the importance sampling weight $\weight$, 
the fourth line uses~\eqref{zetabar_ineq}, 
and the last line uses the fact that, 
conditioned on $\lbrace s^j_{x,a}\rbrace_{j=1}^m$, 
the function $\overline\zeta$ is a member of $\G(m)$.

We can now upper bound the RHS. 
Going forward, we refer to $\zeta(\lbrace s^j_{X,A}\rbrace_{j=1}^m)$ as $\zeta(X,A)$ for short. 
Then for $\lambda>0$ we have:
\allowdisplaybreaks
\begin{align}
    \E_{\Prob_\beta}&\left[\exp\left(\lambda\sup_{\zeta \in \G(m)} \left( \frac{1}{n}\sum_{i=1}^n w(A_i, X_i)\zeta(X_i,A_i) -  \frac{1}{n}\sum_{i=1}^n \E_{\Prob_\beta}\left[ w(A, X_i)\zeta(X_i, A) \Big\vert X_i \right]\right) \right) \right]\nonumber\\
    &\quad\quad\quad= \E_{\Prob_\beta}\left[\exp\left(\lambda\sup_{\zeta \in \G(m)} \left( \frac{1}{n}\sum_{i=1}^n \E_{\Prob_\beta}
    \left[w(A_i, X_i)\zeta(X_i, A_i) - w(A'_i, X_i)\zeta(X_i, A'_i) \Big\vert \lbrace X_i, A_i \rbrace_{i=1}^n \right]\right) \right) \right]\nonumber\\
    &\quad\quad\quad\leq\E_{\Prob_\beta}\left[\exp\left(\lambda\E_{\Prob_\beta}\left[\sup_{\zeta \in \G(m)} \frac{1}{n}\sum_{i=1}^n \left(w(A_i, X_i)\zeta(X_i, A_i) - w(A'_i,X_i)\zeta(X_i, A'_i)\right) \Big\vert \lbrace X_i, A_i \rbrace_{i=1}^n \right]\right) \right]\nonumber\\
    &\quad\quad\quad\leq\E_{\Prob_\beta}\left[\exp\left(\lambda\sup_{\zeta \in \G(m)} \frac{1}{n}\sum_{i=1}^n \left(w(A_i, X_i)\zeta(X_i, A_i) - w(A'_i, X_i)\zeta(X_i, A'_i)\right)\right) \right]\nonumber\\
    &\quad\quad\quad\leq\E_{\Prob_\beta, \R}\left[\exp\left(2\lambda\sup_{\zeta \in \G(m)} \frac{1}{n}\sum_{i=1}^n \xi_i w(A_i, X_i)\zeta(X_i, A_i)\right) \right]\nonumber\\
    &\quad\quad\quad=\E_{\Prob_\beta, \R}\left[\sup_{\zeta \in \G(m)}\exp\left(2\lambda \frac{1}{n}\sum_{i=1}^n \xi_i w(A_i, X_i) \zeta(X_i, A_i)\right) \right]\nonumber \\
    &\quad\quad\quad=\E_{\Prob_\beta, \R}\left[\sup_{t, \varrho}\exp\left(2\lambda \frac{\varrho}{nm}\sum_{j=1}^m \sum_{i=1}^n \xi_i w(A_i, X_i) \mathbbm{1}_{\lbrace s_{X_i,A_i}^j \leq t \rbrace }\right) \right]\label{eq:part1_rhs}
\end{align}
where $\{A'\}_i^n$ are the ghost variables, the second to last inequality uses symmetrization (Lemma~\ref{lem:g_symmetrization}), and the last line uses the definition of $\zeta(X_i,A_i) = \zeta(s^1_{X_i,A_i},...,s^m_{X_i,A_i})$. 

Now, for each $j$, permute the indices $i$ such that $s^{j}_{X_{j(1)},A_{j(1)}} \leq ... \leq s^{j}_{X_{j(i)},A_{j(i)}} \leq ... \leq s^{j}_{X_{j(n)},A_{j(n)}}$. 
Then, for a given $j$, consider the function 
\[
\sum_{i=1}^n \xi_{j(i)} w(A_{j(i)}, X_{j(i)})\mathbbm{1}_{\lbrace s^{j}_{X_{j(i)},A_{j(i)}} \leq t \rbrace},\] 
which equals
\begin{enumerate}
    \item 0 if $t < s^{j}_{X_{j(1)},A_{j(1)}} $, 
    \item $\varrho \sum_{i=1}^k w(A_{j(i)}, X_{j(i)})\xi_{j(i)}$ if there exists $k \in \{1,...,n-1\}$ such that $s^{j}_{X_{j(k)},A_{j(k)}} \leq t \leq s^{j}_{X_{j(k)+1},A_{j(k)+1}}$,
    \item $\varrho \sum_{i=1}^n w(A_{j(i)}, X_{j(i)})\xi_{j(i)}$ otherwise.
\end{enumerate}
Then the RHS of~\eqref{eq:part1_rhs} equals
\begin{align*}
    &\E_{\Prob_\beta, \R}\left[\sup_{t, \varrho}\exp\left(2\lambda \frac{\varrho}{nm}\sum_{j=1}^m \sum_{i=1}^n \xi_{i} w(A_{i}, X_{i}) \mathbbm{1}_{\lbrace s^{j}_{X_i,A_i} \leq t \rbrace }\right) \right] \\
    &\quad\quad= \E_{\Prob_\beta, \R}\left[\max_{k, \varrho}\exp\left(2\lambda \frac{\varrho}{nm}\sum_{j=1}^m \sum_{i=1}^k \xi_{j(i)} w(A_{j(i)}, X_{j(i)}) \right) \right] \\
    &\quad\quad\leq \E_{\Prob_\beta, \R}\left[\max_{j,k, \varrho}\exp\left(2\lambda \frac{\varrho}{n} \sum_{i=1}^k \xi_{j(i)} w(A_{j(i)}, X_{j(i)}) \right) \right]. 
\end{align*}

Further, we have that 
\begin{align*}
    &\max_{j,k, \varrho}\exp\left(2\lambda \frac{\varrho}{n} \sum_{i=1}^k \xi_{j(i)} w(A_{j(i)}, X_{j(i)}) \right)\\
    &= \max_{j,k}\bigg(\exp\left(\frac{2\lambda}{n}\sum_{i}^k w(A_{j(i)}, X_{j(i)})\xi_{j(i)}\right)\Ind_{\lbrace \sum_i^k w(A_{j(i)}, X_{j(i)})\xi_{j(i)} \geq 0 \rbrace} \\
    &\quad\quad\quad\quad+ \exp\left(-\frac{2\lambda}{n}\sum_{i}^k w(A_{j(i)}, X_{j(i)})\xi_{j(i)}\right)\Ind_{\lbrace \sum_i^k w(a_{j(i)}, x_{j(i)})\xi_{j(i)} < 0 \rbrace} \bigg) \\ 
    &\leq 2\max_{j,k}\exp\left(\frac{2\lambda}{n}\sum_{i}^k w(A_{j(i)}, X_{j(i)})\xi_{j(i)}\right)\Ind_{\lbrace \sum_i^k w(A_{j(i)}, X_{j(i)})\xi_{j(i)} \geq 0 \rbrace}.
\end{align*}
Putting it together, we have that 
\begin{align}
    &\E_{\Prob_\beta}\left[\exp \left( \lambda \sup_{\zeta \in \G(m)} \left\vert \frac{1}{n}\sum_{i=1}^n w(A_i, X_i)\zeta(X_i,A_i) - \frac{1}{n}\sum_{i=1}^n \E_{A \sim \pi(\cdot|X_i)}\left[\zeta(X_i, A) | X_i\right] \right\vert \right)\right] \nonumber\\
    &\quad\quad\quad\quad\leq 2\E_{\Prob_\beta, \R}\left[ \max_{j,k}\exp\left(\frac{2\lambda}{n}\sum_{i}^k w(A_{j(i)}, X_{j(i)})\xi_{j(i)}\right)\Ind_{\lbrace \sum_i^k w(A_{j(i)}, X_{j(i)})\xi_{j(i)} \geq 0 \rbrace} \right] \label{eq:dr_other_ineq}
\end{align}
Now we are left to bound the RHS of~\eqref{eq:dr_other_ineq}. Using Lemma~\ref{lemma:CDF_Convex}, 
\begin{align*}
    &\E_{\Prob_\beta, \R}\left[ \max_{j,k}\exp\left(\frac{2\lambda}{n}\sum_{i}^k w(A_{j(i)}, X_{j(i)})\xi_{j(i)}\right)\Ind_{\lbrace \max_k\sum_i^k w(A_{j(i)}, X_{j(i)})\xi_{j(i)} \geq 0 \rbrace} \right] \\
    &\quad\quad\leq \Prob_\beta\left( \max_k \frac{2\lambda}{n} \sum_{i}^k w(A_{j(i)}, X_{j(i)}) \xi_{j(i)} \geq 0 \right) + 2\lambda \sum_j \int_{0}^\infty \exp(\lambda t)\Prob\left(\frac{2\lambda}{n}\sum_{i=1}^n  w(A_{j(i)}, X_{j(i)})\xi_{j(i)} \geq t \right) dt. 
\end{align*}
Similarly, for any $j$, we have 
\begin{align*}
    &\E_{\Prob_\beta, \R}\left[ \exp\left(\frac{2\lambda}{n}\sum_{i}^n w(A_{j(i)}, X_{j(i)})\xi_{j(i)}\right)\Ind_{\lbrace \sum_i^k w(A_{j(i)}, X_{j(i)})\xi_{j(i)} \geq 0 \rbrace} \right] \\
    &\quad\quad= \Prob_\beta\left(\frac{2\lambda}{n}\sum_i^n w(A_{j(i)}, X_{j(i)})\xi_{j(i)} \geq 0 \right) + \lambda\int_{0}^\infty\exp(\lambda t) \Prob\left( \frac{2\lambda}{n}\sum_{i=1}^n w(A_{j(i)}, X_{j(i)})\xi_{j(i)}\geq t \right) dt
\end{align*}
Putting these two together, we have 
\begin{align*}
    &\E_{\Prob_\beta, \R}\left[ \max_{j,k}\exp\left(\frac{2\lambda}{n}\sum_{i}^k w(A_{j(i)}, X_{j(i)})\xi_{j(i)}\right)\Ind_{\lbrace \sum_i^k w(A_{j(i)}, X_{j(i)})\xi_{j(i)} \geq 0 \rbrace} \right] \\
    &\quad\quad\leq \sum_j\Prob_\beta\left( \max_k \frac{2\lambda}{n} \sum_{i}^k w(A_{j(i)}, X_{j(i)}) \xi_{j(i)} \geq 0 \right) - 2\sum_j\Prob_\beta\left(\frac{2\lambda}{n}\sum_i^n w(A_{j(i)}, X_{j(i)})\xi_{j(i)} \geq 0 \right) \\ 
    &\quad\quad\quad+2\sum_j\E_{\Prob_\beta, \R}\left[ \exp\left(\frac{2\lambda}{n}\sum_{i}^n w(A_{j(i)}, X_{j(i)})\xi_{j(i)}\right)\Ind_{\lbrace \sum_i^n w(A_{j(i)}, X_{j(i)})\xi_{j(i)} \geq 0 \rbrace} \right] \\
    &\quad\quad\leq 2\sum_j \E_{\Prob_\beta, \R}\left[ \exp\left(\frac{2\lambda}{n}\sum_{i}^n w(A_{j(i)}, X_{j(i)})\xi_{j(i)}\right)\Ind_{\lbrace \sum_i^n w(A_{j(i)}, X_{j(i)})\xi_{j(i)} \geq 0 \rbrace} \right] \\
    &\quad\quad\leq 2m \E_{\Prob_\beta, \R}\left[ \exp\left(\frac{2\lambda}{n}\sum_{i}^n w(A_{j(i)}, X_{j(i)})\xi_{j(i)}\right)\right] \\ 
    &\quad\quad\leq 2m\exp\left(\frac{2\lambda^2 w_{max}^2}{n}\right)
\end{align*}
where the last inequality uses the fact that $\xi$ is a Rademacher random variable, and $w(A,X) \leq w_{max}$. 
Finally, using Markov's inequality, 
\begin{align*}
    &\Prob_\beta\left(  \sup_t \left\vert \frac{1}{n}\sum_{i=1}^{n}\overline{G}(t; X_i, \pi) - w(A_i, X_i)\overline{G}(t; X_i, A_i) \right\vert \geq \epsilon + \frac{1}{m} \right) \\
    &\quad\quad\leq \Prob_\beta\left( \exp\left( \lambda \sup_t \left\vert \frac{1}{n}\sum_{i=1}^n w(A_i, X_i)\zeta(X_i,A_i) - \frac{1}{n}\sum_{i=1}^n \E_{\Prob_\beta}\left[\zeta(X_i, A) | X_i\right]  \right\vert \right) \geq \exp(\lambda\epsilon) \right)\\
    &\quad\quad\leq 4m\exp\left(\frac{2\lambda^2 w_{max}^2}{n} - \lambda\epsilon  \right)
\end{align*}

Because this holds for any $\lambda > 0$, we can minimize the RHS over $\lambda$: 
\begin{align*}
    \Prob_\beta\left(  \sup_t \left\vert \frac{1}{n}\sum_{i=1}^{n}\overline{G}(t;X_i, \pi) - w(A_i, X_i)\overline{G}(t; X_i, A_i) \right\vert \geq \epsilon + \frac{1}{m} \right) &\leq \inf_{\lambda > 0} 4m\exp\left(\frac{2\lambda^2 w_{max}^2}{n} - \lambda\epsilon  \right)\\
    &= 4m\exp\left( \frac{-n\epsilon }{8w_{max}^2}\right).
\end{align*}
Then we have 
\begin{align*}
    \Prob_\beta\left(  \sup_t \left\vert \frac{1}{n}\sum_{i=1}^{n}\overline{G}(t; X_i, \pi) - w(A_i, X_i)\overline{G}(t; X_i, A_i) \right\vert \geq \sqrt{\frac{8w_{max}^2}{n}\log\frac{4m}{\delta}} + \frac{1}{m} \right) &\leq \delta.
\end{align*}
Setting $m = \sqrt{n/8w_{max}^2}$ gives the theorem statement: 
\begin{align*}
    \Prob_\beta\left(  \sup_t \left\vert \frac{1}{n}\sum_{i=1}^{n}\overline{G}(t; X_i, \pi) - w(A_i, X_i)\overline{G}(t; X_i, A_i) \right\vert \geq \sqrt{\frac{32w_{max}^2}{n}\log\frac{(2n)^{1/2}}{ w_{max}\delta}} \right) &\leq \delta.
\end{align*}
\end{proof}

\subsubsection*{Auxiliary Lemmas}
\begin{lemma}\label{Lemma:approximation}
For any $\zeta$, a non-decreasing function with support $[0,D]$,
there exists $m$ points $s^1....s^m \in \Q^m$ 
such that 
for a function of the form, 

\[
    \overline\zeta(t; s^1,...,s^m) = \frac{1}{m}\sum_{j=1}^{m}\Ind_{\lbrace s^j \leq t \rbrace}, \;\;\forall t \in \Real
\]
the following inequality holds:
\begin{align*}
    \|\zeta-\wb \zeta\|_{\infty}\leq \frac{1}{2m}.
\end{align*}
\end{lemma}

\begin{proof}[Proof of Lemma~\ref{Lemma:approximation}]

Uniformly partition the interval $[0,D]$ to $m$ partitions, 
with partition points $\{\frac{j}{D}\}_{j=0}^{m}$.
We construct the set $\lbrace s_j \rbrace_{j=1}^m$ 
using the following procedure. 
For any $j\in\{1,\ldots,m\}$ and the corresponding partition point $\frac{j-1}{D}$, let $s^j\in\Q$ be a point such that either $\lim_{t\rightarrow{s^j_-}}\zeta(t)=\frac{j-1}{m}+\frac{1}{2m}$ or $\lim_{t\rightarrow{s^j_+}}\zeta(t)=\frac{j-1}{m}+\frac{1}{2m}$ (e.g., as illustrated in Figure~\ref{fig:approximation}). Then for any $t$, $\wb\zeta(t)$ is $\frac{1}{2m}$-close to $\zeta(t)$.
\begin{figure}[h]
    \centering
    \includegraphics[scale=0.4, trim={0cm 2cm 0 5cm},clip]{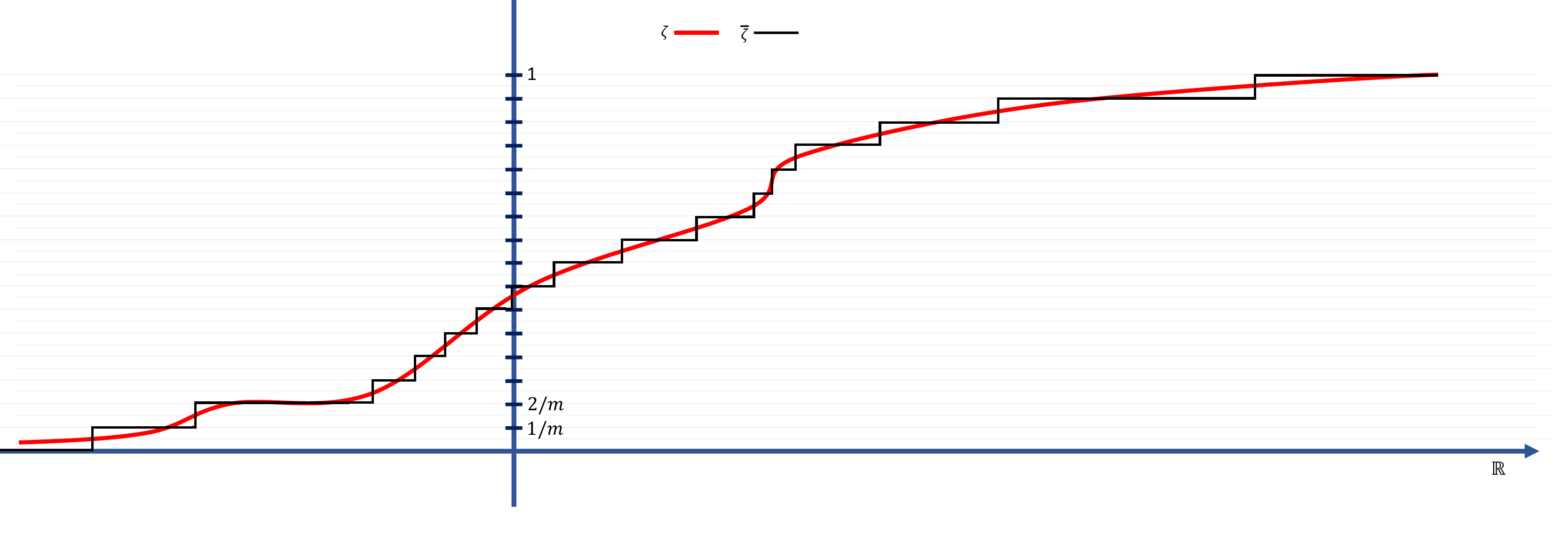}
    \caption{Approximating monotonic function $\zeta$ with $\wb\zeta$. }
    \label{fig:approximation}
\end{figure}
\end{proof}

\begin{lemma}\label{lem:g_symmetrization}
For the function class $\G$ defined in Appendix~\ref{appendix:dr_error}, we have for any $\lambda > 0$ that
    \begin{align*}
        &\E_{\Prob_\beta}\left[\exp\left(\lambda\sup_{\zeta \in \G(m)} \frac{1}{n}\sum_{i=1}^n \left(w(A_i, X_i)\zeta(X_i, A_i) - w(A'_i, X_i)\zeta(X_i, A'_i)\right)\right) \right] \\
        &\quad\quad\leq \E_{\Prob_\beta, \R}\left[\sup_{\zeta \in \G(m)} \exp\left(2\lambda \frac{1}{n}\sum_{i=1}^n \xi_i w(A_i, X_i)\zeta(X_i, A_i)\right) \right]
    \end{align*}
where contexts and actions $X, A, A' \sim \Prob_\beta$, and Rademacher random variables $\xi_i \sim \R$. 
\end{lemma}
\begin{proof}
    For each $i = 1, ..., n$, and let $\xi_i$ be i.i.d. Rademacher random variables. Set 
    \[
        A_i^+ = 
        \begin{cases}
            A_i, & \text{if } \xi_i = 1\\
            A_i', & \text{if } \xi_i = -1
        \end{cases}
    \]
    \[
        A_i^- = 
        \begin{cases}
            A_i', & \text{if } \xi_i = 1\\
            A_i, & \text{if } \xi_i = -1
        \end{cases}
    \]
    We have that, conditioned on $X_i$, $(A_i^+, A_i^-) \stackrel{d}{=} (A_i, A'_i)$. Then 
    \begin{align*}
        &\E_{\Prob_\beta}\left[\exp\left(\lambda\sup_{\zeta \in \G(m)} \frac{1}{n}\sum_{i=1}^n \left(w(A_i, X_i)\zeta(X_i, A_i) - w(A'_i, X_i)\zeta(X_i, A'_i)\right)\right) \right] \\ 
        &\quad\quad= \E_{\Prob_\beta}\left[\exp\left(\lambda\sup_{\zeta \in \G(m)} \frac{1}{n}\sum_{i=1}^n \left(w(A_i^+, X_i)\zeta(X_i, A_i^+) - w(A_i^-, X_i)\zeta(X_i, A_i^-)\right)\right) \right] \\ 
        &\quad\quad= \E_{\Prob_\beta, \R}\left[\exp\left(\lambda\sup_{\zeta \in \G(m)} \frac{1}{n}\sum_{i=1}^n \xi_i \left(w(A_i, X_i)\zeta(X_i, A_i) - w(A'_i, X_i)\zeta(X_i, A'_i)\right)\right) \right] \\ 
    \end{align*}
    Our last step is to bound the last line of the above display. 
    \begin{align*}
        &= \E_{\Prob_\beta,\R}\left[\exp\left(\lambda\sup_{\zeta \in \G(m)} \frac{1}{2}\left(\frac{2}{n}\sum_{i=1}^n \xi_i w(A_i, X_i)\zeta(X_i, A_i) - \frac{2}{n}\sum_{i=1}^n \xi_i w(A'_i, X_i)\zeta(X_i, A'_i)\right)\right) \right] \\
        &\leq \frac{1}{2}\E_{\Prob_\beta,\R}\left[\exp\left(\lambda\sup_{\zeta \in \G(m)} \frac{2}{n}\sum_{i=1}^n \xi_i w(A_i, X_i)\zeta(X_i, A_i)\right) \right] \\
        &\quad\quad+ \frac{1}{2} \E_{\Prob_\beta,\R}\left[\exp\left(\lambda\sup_{\zeta \in \G(m)} \frac{2}{n}\sum_{i=1}^n (-\xi_i)w(A'_i, X_i)\zeta(X_i, A'_i)\right) \right] \\
        &= \E_{\Prob_\beta,\R}\left[\exp\left(\lambda\sup_{\zeta \in \G(m)} \frac{2}{n}\sum_{i=1}^n \xi_i w(A_i, X_i)\zeta(X_i, A_i)\right) \right]
    \end{align*}
\end{proof}

\begin{lemma}\label{lem:x_concentration}
Let $G(t; X, \pi) = \E_{\Prob}[\mathbbm{1}_{\lbrace R \leq t \rbrace} | X]$ be the conditional \CDF of returns for all $x \in \X$. Then for $\delta \in (0, 1]$, 
\begin{align*}
    \Prob_\beta\left(  \sup_t \left\vert \frac{1}{n}\sum_{i=1}^{n}G(t; X_i, \pi) - F(t) \right\vert \geq \sqrt{\frac{32}{n}\log\frac{(2n)^{1/2}}{ \delta}} \right) &\leq \delta.
\end{align*}
\end{lemma}

\begin{proof}
Since $G$ is a valid \CDF,  
we apply Lemma \ref{Lemma:approximation} to $\overline{G}$.  
Consider a function of the form 
\[
 \overline\zeta(t;s^1, ..., s^m) = \frac{1}{m}\sum_{j=1}^m \mathbbm{1}_{\lbrace s^i \leq t \rbrace} 
\]
The function $\overline\zeta$ can be seen as a stepwise CDF function, where each step is $1/m$ and occurs at points $\lbrace s^j \rbrace_{j=1}^m$. 

Lemma~\ref{Lemma:approximation} approximates $\overline{G}$ using such $1/m$-stepwise \CDF{}s. 
For each context $x$, 
let $s^1_{x}, ..., s^m_{x} \in \mathbb{Q}^m$ be the points chosen according to the deterministic procedure in Lemma \ref{Lemma:approximation}, 
such that the following inequality holds:
\begin{align}
    \sup_t \left\vert G(t; x, \pi) - \overline\zeta\left(t;\lbrace s^j_{x}\rbrace_{j=1}^m\right) \right\vert \leq \frac{1}{2m}. 
\end{align}

Next, consider the class of functions \begin{align*}
    \G(m) :=\Big\lbrace \zeta(s^1,...,s^m):=\frac{1}{m}\varrho\sum_{j=1}^{m}\Ind_{\lbrace s^j \leq t \rbrace}: \forall t\in\Real, \varrho \in \lbrace -1, +1 \rbrace ; \lbrace s^j\rbrace_{j=1}^m\in\Q^m\Big\rbrace
\end{align*}
Note that, $\overline\zeta$ is a subset of the function class $\G(m)$, e.g. $\overline\zeta \left(t; \lbrace s^j_{x}\rbrace_{j=1}^m\right) \in \G(m)$. 

Then our problem becomes 
\allowdisplaybreaks

\begin{align*}
    &\sup_t \left\vert \frac{1}{n} \sum_{i=1}^n G(t; X_i, \pi) -  F(t) \right\vert\\ 
    &\quad\quad= \sup_{t} \left\vert \frac{1}{n} \sum_{i=1}^n G(t; X_i, \pi) -  \E_{\Prob_\beta}\left[\frac{1}{n}\sum_{i=1}^{n}G(t; X, \pi)\right] \right\vert\\
    &\quad\quad\leq \sup_{t} \left\vert \frac{1}{n}\sum_{i=1}^n \overline\zeta\left(t;\lbrace s^j_{X_i}\rbrace_{j=1}^m\right) - \E_{\Prob_\beta}\left[\frac{1}{n}\sum_{i=1}^{n}\overline\zeta\left(t;\lbrace s^j_{X_i}\rbrace_{j=1}^m\right)\right] \right\vert + \frac{1}{m}\\
    &\quad\quad\leq \sup_{\zeta \in \G(m)} \left\vert \frac{1}{n}\sum_{i=1}^n \zeta\left(\lbrace s^j_{X_i}\rbrace_{j=1}^m\right) - \E_{\Prob_\beta}\left[\frac{1}{n}\sum_{i=1}^{n}\zeta\left(\lbrace s^j_{X_i}\rbrace_{j=1}^m\right)\right] \right\vert + \frac{1}{m}\\
\end{align*}
 
We can now upper bound the RHS. 
Going forward, we refer to $\zeta(\lbrace s^j_{X}\rbrace_{j=1}^m)$ as $\zeta(X)$ for short. 
Then for $\lambda>0$ we have:
\allowdisplaybreaks
\begin{align}
    \E_{\Prob_\beta}&\left[\exp\left(\lambda\sup_{\zeta \in \G(m)} \left( \frac{1}{n}\sum_{i=1}^n \zeta(X_i) -  \E_{\Prob_\beta}\left[ \frac{1}{n}\sum_{i=1}^n \zeta(X_i) \right]\right) \right) \right]\nonumber\\
    &\quad\quad\quad= \E_{\Prob_\beta}\left[\exp\left(\lambda\sup_{\zeta \in \G(m)} \left( \frac{1}{n}\sum_{i=1}^n \E_{\Prob_\beta}
    \left[\zeta(X_i) - \zeta(X'_i) \Big\vert \lbrace X_i \rbrace_{i=1}^n \right]\right) \right) \right]\nonumber\\
    &\quad\quad\quad\leq\E_{\Prob_\beta}\left[\exp\left(\lambda\E_{\Prob_\beta}\left[\sup_{\zeta \in \G(m)} \frac{1}{n}\sum_{i=1}^n \left(\zeta(X_i) - \zeta(X'_i)\right) \Big\vert \lbrace X_i \rbrace_{i=1}^n \right]\right) \right]\nonumber\\
    &\quad\quad\quad\leq\E_{\Prob_\beta}\left[\exp\left(\lambda\sup_{\zeta \in \G(m)} \frac{1}{n}\sum_{i=1}^n \left(\zeta(X_i) - \zeta(X'_i, A'_i)\right)\right) \right]\nonumber\\
    &\quad\quad\quad=\E_{\Prob_\beta, \R}\left[\exp\left(\lambda\sup_{\zeta \in \G(m)} \frac{1}{n}\sum_{i=1}^n \xi_i\left(\zeta(X_i) - \zeta(X'_i)\right)\right) \right]\nonumber\\
    &\quad\quad\quad\leq\E_{\Prob_\beta, \R}\left[\exp\left(2\lambda\sup_{\zeta \in \G(m)} \frac{1}{n}\sum_{i=1}^n \xi_i \zeta(X_i)\right) \right]\nonumber\\
    &\quad\quad\quad=\E_{\Prob_\beta, \R}\left[\sup_{\zeta \in \G(m)}\exp\left(2\lambda \frac{1}{n}\sum_{i=1}^n \xi_i \zeta(X_i)\right) \right]\nonumber \\
    &\quad\quad\quad=\E_{\Prob_\beta, \R}\left[\sup_{t, \varrho}\exp\left(2\lambda \frac{\varrho}{nm}\sum_{j=1}^m \sum_{i=1}^n \xi_i\mathbbm{1}_{\lbrace s_{X_i}^j \leq t \rbrace }\right) \right]\label{eq:part1_rh}
\end{align}
where $\{X'\}_i^n$ are the ghost variables, and the last line uses the definition of $\zeta(X_i) = \zeta(s^1_{X_i},...,s^m_{X_i})$. 

Now, for each $j$, permute the indices $i$ such that $s^{j}_{X_{j(1)}} \leq ... \leq s^{j}_{X_{j(i)}} \leq ... \leq s^{j}_{X_{j(n)}}$. 
Then, for a given $j$, consider the function 
\[
\sum_{i=1}^n \xi_{j(i)} \mathbbm{1}_{\lbrace s^{j}_{X_{j(i)}} \leq t \rbrace},\] 
which equals
\begin{enumerate}
    \item 0 if $t < s^{j}_{X_{j(1)}} $, 
    \item $\varrho \sum_{i=1}^k \xi_{j(i)}$ if there exists $k \in \{1,...,n-1\}$ such that $s^{j}_{X_{j(k)}} \leq t \leq s^{j}_{X_{j(k)+1}}$,
    \item $\varrho \sum_{i=1}^n \xi_{j(i)}$ otherwise.
\end{enumerate}
Then the RHS of~\eqref{eq:part1_rh} equals
\begin{align*}
    &\E_{\Prob_\beta, \R}\left[\sup_{t, \varrho}\exp\left(2\lambda \frac{\varrho}{nm}\sum_{j=1}^m \sum_{i=1}^n \xi_{i}  \mathbbm{1}_{\lbrace s^{j}_{X_i} \leq t \rbrace }\right) \right] \\
    &\quad\quad= \E_{\Prob_\beta, \R}\left[\max_{k, \varrho}\exp\left(2\lambda \frac{\varrho}{nm}\sum_{j=1}^m \sum_{i=1}^k \xi_{j(i)} \right) \right] \\
    &\quad\quad\leq \E_{\Prob_\beta, \R}\left[\max_{j,k, \varrho}\exp\left(2\lambda \frac{\varrho}{n} \sum_{i=1}^k \xi_{j(i)}  \right) \right]. 
\end{align*}

Further, we have that 
\begin{align*}
    &\max_{j,k, \varrho}\exp\left(2\lambda \frac{\varrho}{n} \sum_{i=1}^k \xi_{j(i)}\right)\\
    &= \max_{j,k}\bigg(\exp\left(\frac{2\lambda}{n}\sum_{i}^k \xi_{j(i)}\right)\Ind_{\lbrace \sum_i^k \xi_{j(i)} \geq 0 \rbrace} + \exp\left(-\frac{2\lambda}{n}\sum_{i}^k \xi_{j(i)}\right)\Ind_{\lbrace \sum_i^k \xi_{j(i)} < 0 \rbrace} \bigg) \\ 
    &\leq 2\max_{j,k}\exp\left(\frac{2\lambda}{n}\sum_{i}^k \xi_{j(i)}\right)\Ind_{\lbrace \sum_i^k \xi_{j(i)} \geq 0 \rbrace}.
\end{align*}
Putting it together, we have that 
\begin{align}
    &\E_{\Prob_\beta}\left[\exp \left( \lambda \sup_{\zeta \in \G(m)} \left\vert \frac{1}{n}\sum_{i=1}^n \zeta(X_i) - \E_{\Prob_\beta}\left[\frac{1}{n}\sum_{i=1}^n \zeta(X_i)\right] \right\vert \right)\right] \nonumber\\
    &\quad\quad\quad\quad\leq 2\E_{\Prob_\beta, \R}\left[ \max_{j,k}\exp\left(\frac{2\lambda}{n}\sum_{i}^k\xi_{j(i)}\right)\Ind_{\lbrace \sum_i^k \xi_{j(i)} \geq 0 \rbrace} \right] \label{eq:dr_other_ineq2}
\end{align}
Now we are left to bound the RHS of~\eqref{eq:dr_other_ineq2}. Using Lemma~\ref{lemma:CDF_Convex}, 
\begin{align*}
    &\E_{\Prob_\beta, \R}\left[ \max_{j,k}\exp\left(\frac{2\lambda}{n}\sum_{i}^k\xi_{j(i)}\right)\Ind_{\lbrace \max_k\sum_i^k \xi_{j(i)} \geq 0 \rbrace} \right] \\
    &\quad\quad\leq \Prob_\beta\left( \max_k \frac{2\lambda}{n} \sum_{i}^k  \xi_{j(i)} \geq 0 \right) + 2\lambda \sum_j \int_{0}^\infty \exp(\lambda t)\Prob\left(\frac{2\lambda}{n}\sum_{i=1}^n \xi_{j(i)} \geq t \right) dt. 
\end{align*}
Similarly, for any $j$, we have 
\begin{align*}
    &\E_{\Prob_\beta, \R}\left[ \exp\left(\frac{2\lambda}{n}\sum_{i}^n\xi_{j(i)}\right)\Ind_{\lbrace \sum_i^k\xi_{j(i)} \geq 0 \rbrace} \right] \\
    &\quad\quad= \Prob_\beta\left(\frac{2\lambda}{n}\sum_i^n w(A_{j(i)}, X_{j(i)})\xi_{j(i)} \geq 0 \right) + \lambda\int_{0}^\infty\exp(\lambda t) \Prob\left( \frac{2\lambda}{n}\sum_{i=1}^n \xi_{j(i)}\geq t \right) dt
\end{align*}
Putting these two together, we have 
\begin{align*}
    &\E_{\Prob_\beta, \R}\left[ \max_{j,k}\exp\left(\frac{2\lambda}{n}\sum_{i}^k \xi_{j(i)}\right)\Ind_{\lbrace \sum_i^k \xi_{j(i)} \geq 0 \rbrace} \right] \\
    &\quad\quad\leq \sum_j\Prob_\beta\left( \max_k \frac{2\lambda}{n} \sum_{i}^k  \xi_{j(i)} \geq 0 \right) - 2\sum_j\Prob_\beta\left(\frac{2\lambda}{n}\sum_i^n \xi_{j(i)} \geq 0 \right) \\ 
    &\quad\quad\quad+2\sum_j\E_{\Prob_\beta, \R}\left[ \exp\left(\frac{2\lambda}{n}\sum_{i}^n \xi_{j(i)}\right)\Ind_{\lbrace \sum_i^n \xi_{j(i)} \geq 0 \rbrace} \right] \\
    &\quad\quad\leq 2\sum_j \E_{\Prob_\beta, \R}\left[ \exp\left(\frac{2\lambda}{n}\sum_{i}^n \xi_{j(i)}\right)\Ind_{\lbrace \sum_i^n \xi_{j(i)} \geq 0 \rbrace} \right] \\
    &\quad\quad\leq 2m \E_{\Prob_\beta, \R}\left[ \exp\left(\frac{2\lambda}{n}\sum_{i}^n \xi_{j(i)}\right)\right] \\ 
    &\quad\quad\leq 2m\exp\left(\frac{2\lambda^2}{n}\right)
\end{align*}
where the last inequality uses the fact that $\xi$ is a Rademacher random variable. 
Finally, using Markov's inequality, 
\begin{align*}
    &\Prob_\beta\left(  \sup_t \left\vert \frac{1}{n}\sum_{i=1}^{n}G(t; X_i, \pi) - F(t)\right\vert \geq \epsilon + \frac{1}{m} \right) \\
    &\quad\quad\leq \Prob_\beta\left( \exp\left( \lambda \sup_t \left\vert \frac{1}{n}\sum_{i=1}^n \zeta(X_i) - \E_{\Prob_\beta}\left[\frac{1}{n}\sum_{i=1}^n \zeta(X_i)\right]  \right\vert \right) \geq \exp(\lambda\epsilon) \right)\\
    &\quad\quad\leq 4m\exp\left(\frac{2\lambda^2}{n} - \lambda\epsilon  \right)
\end{align*}

Because this holds for any $\lambda > 0$, we can minimize the RHS over $\lambda$: 
\begin{align*}
    \Prob_\beta\left(  \sup_t \left\vert \frac{1}{n}\sum_{i=1}^{n}G(t;X_i, \pi) - F(t) \right\vert \geq \epsilon + \frac{1}{m} \right) &\leq \inf_{\lambda > 0} 4m\exp\left(\frac{2\lambda^2 }{n} - \lambda\epsilon  \right)\\
    &= 4m\exp\left( \frac{-n\epsilon }{8}\right).
\end{align*}
Then we have 
\begin{align*}
    \Prob_\beta\left(  \sup_t \left\vert \frac{1}{n}\sum_{i=1}^{n}G(t; X_i, \pi) - F(t) \right\vert \geq \sqrt{\frac{8}{n}\log\frac{4m}{\delta}} + \frac{1}{m} \right) &\leq \delta.
\end{align*}
Setting $m = \sqrt{n/8}$ gives the theorem statement: 
\begin{align*}
    \Prob_\beta\left(  \sup_t \left\vert \frac{1}{n}\sum_{i=1}^{n}G(t; X_i, \pi) - F(t) \right\vert \geq \sqrt{\frac{32}{n}\log\frac{(2n)^{1/2}}{\delta}} \right) &\leq \delta.
\end{align*}
\end{proof}

\section{Proofs for Risk Functional Estimation (Section~\ref{sec:risk})}
\begin{proof}[Proof of Theorem~\ref{thm:confidence_general}]
By the definition of $L$-Lipschitz risk functionals, for the \CDF{}s $F$ and $\wh{F}$, 
\begin{align*}
    \vert \rho(\wh{F}) - \rho(F) \vert  &\leq L \Vert \wh{F} - F \Vert_\infty \\ 
    &\leq L\epsilon
\end{align*}
with probability at least $1-\delta$, where the last line uses the fact that $\wh{F}$ is $\epsilon$-close to $F$ with probability at least $1-\delta$. 
\end{proof}

\clearpage
\section{Risk Estimation with Unknown Behavior Policy}\label{appendix:estimated_policy}
We begin this section with a consideration of estimators when the behavior policy is unknown, and must be modeled or estimated, which we call $\wh\beta$. 
We first define the IS, DR, and DI estimators using $\wh\beta$, then derive their bias and variance expressions. 
To differentiate between the estimator that use $\beta$ and the estimators that use $\wh\beta$, we call the latter $\wt{F}$ while continuing to call the former $\wh{F}$.  

The proofs of bias and variance begins with derivations for the DR estimator with estimated policy, from which the bias and variance of the remaining estimators can be derived as special cases. 

Let $\wh\beta$ be the estimated behavior policy, and let $\wh\weight(a,x) := \frac{\pi(a|x)}{\wh\beta(a|x)}$ be the importance weight with estimated policy. Then the importance sampling (IS) estimator is given by
\begin{equation}\label{eq:is_policy_est}
    \wt{F}_{\text{IS}}(t) := \frac{1}{n}\sum_{i=1}^n \wh\weight(a_i,x_i) \Ind_{\{r_i \leq t\}}
\end{equation}
Then doubly robust (DR) estimator is: 
\begin{equation}\label{eq:dr_policy_est}
    \wt{F}_{\text{DR}}(t) := \frac{1}{n}\sum_{i=1}^n \wh\weight(a_i,x_i) \left(\Ind_{\{r_i \leq t\}} - \overline{G}(t;x_i, a_i)\right) + \overline{G}(t;x_i, \pi)
\end{equation}
And the direct method (DI) estimator is still defined to be 
\begin{equation}\label{eq:di_policy_est}
    \wh{F}_{\text{DI}}(t) := \frac{1}{n}\sum_{i=1}^n \overline{G}(t;x_i, \pi)
\end{equation}
Note that the direct estimator does not depend on the behavior policy, and thus we continue to call it $\wt{F}_{\text{DI}}$.  

\subsection{Bias and Variance}
Next, we analyze the bias and variance of these estimators. 
Define $\Delta(a,x,t)$ to be the additive error between $G$ and the model $\overline{G}$, and define $\delta(x,a)$ to be the multiplicative error of the estimate $\wh\beta$, that is: 
\begin{align*}
    \Delta(t;x,a) &:= \overline{G}(t;x,a) - G(t;x,a), \\
    \delta(x,a) &:= 1 - \beta(a|x) / \wh\beta(a|x). 
\end{align*}
Note that when $\beta$ is known or $\wh\beta = \beta$ for all $x,a$, $\delta(x,a) = 0$,  
The bias of the IS estimator then given in Lemma \ref{lem:is_estim_policy_bias}, in terms of $\delta$ and the conditional reward distribution $G$. 

\begin{lemma}[Bias and Variance of IS Estimator with $\wh\beta$.]\label{lem:is_estim_policy_bias}
    The expectation of the IS estimator is 
    \begin{align*}
        \E_{\mathbb{P}_\beta}[\wt{F}_{\text{IS}}(t)] &= F(t) + \E_{\Prob}[\delta(A,X)G(t;X,\pi)]
    \end{align*}
    When $\wh{\beta}(a|x) = \beta(a|x)$ for all $a,x$, the IS estimator is unbiased and $ \E_{\mathbb{P}_\beta}[\wh{F}_{\text{IS}}(t)] = F(t)$. Further, the variance is 
    \begin{align}
        \Var_{\Prob_\beta}[\wt{F}_{\text{IS}}(t)] &=\frac{1}{n}\E_{\Prob}\left[ \left( 1-\delta(A,X) \right)^2\sigma^2(t;X,A) \right] + \frac{1}{n}\Var_{\Prob}\left[\E_{\Prob}\left[ (1-\delta(A,X))G(t;X,A) \vert X\right] \right] \nonumber\\
        &\quad+ \frac{1}{n}\E_{\Prob}\left[\Var_{\Prob_\beta}\left[ \wh{w}(A,X)G(t;X,A) \vert X \right]\right]\label{is_estim_policy_variance}
    \end{align}
\end{lemma}
The expression for variance is broken down into three terms. The first represents randomness in the rewards, and the second represents variance from the aleatoric uncertainty due to randomness over contexts $X$. The final term represents variance arising from using importance sampling, and is proportional to the true CDF of conditional rewards $G$. 

The following lemma, similarly, derives the bias and variance for the DR estimator:

\begin{lemma}[Bias and Variance of DR Estimator with $\wh\beta$.]\label{lem:dr_estim_policy_bias}
The pointwise expectation of the DR estimator is 
\begin{equation*}
    \E_{\Prob_\beta}[\wt{F}_{\text{DR}}(t)] = F(t) + \E_{\Prob}[\delta(X,A)\Delta(t;X,A)]
\end{equation*}
Further, when there is perfect knowledge of the behavior policy $\beta$, e.g. $\hat{\beta}(a|x) = \beta(a|x)$ for all $a,x$, the DR estimator is unbiased and 
\begin{equation*}
    \E_{\Prob_\beta}[\wt{F}_{\text{DR}}(t)] = F(t)
\end{equation*}
The variance of the doubly robust estimator is given by
\begin{align}
    \Var_{\Prob_\beta}[\wt{F}_{\text{DR}}(t)] &=\frac{1}{n}\E_{\Prob}\left[ \left( 1-\delta(A,X) \right)^2\sigma^2(t;X,A) \right] + \frac{1}{n}\Var_{\Prob}\left[\E_{\Prob}\left[ \delta(A,X)\Delta(t;X,A) + G(t;X,A) \vert X\right] \right] \nonumber \\
    &\quad+ \frac{1}{n}\E_{\Prob}\left[\Var_{\Prob_\beta}\left[ \wh{w}(A,X)\Delta(t;X,A) \vert X \right]\right]\label{dr_estim_policy_variance}
\end{align}
\end{lemma}
Because the DR estimator takes advantage of both policy and reward estimates, 
it is unbiased whenever either the estimated policy or estimated reward is unbiased. 
Further, when we have access to the true behavior policy $\beta$ and $\wh{w} = w$, 
it retains the unbiasedness of the IS estimator. 

Compared to the IS estimator, the DR estimator may also have pointwise reduced variance. When the variances of the IS estimator~\eqref{is_estim_policy_variance} and the DR estimator~\eqref{dr_estim_policy_variance} are compared, 
the first term is identical, and the middle term is of similar magnitude because the randomness in contexts $X$ is endemic.
The third term is the primary difference. 
For the IS estimator, it is proportional to $G$, 
but for the DR estimator, it is proportional to the error $\Delta$ between the estimated conditional CDF $\overline{G}$ and the true $G$. 
Thus, this term can be much larger in the IS estimator when $\wh\weight$ is large and the error $\Delta$ is smaller than $G$. This demonstrates that the DR estimator retains the low bias of the IS estimator, but has the advantage of reduced variance. 

Next, Lemma \ref{lem:di_bias} gives the bias and variance of the DI estimator, which is directly related to the bias and variance of the conditional distribution model $\overline{G}$. 
\begin{lemma}[Bias of DI Estimator with $\wh\beta$.]\label{lem:di_bias}
The bias is 
$$\E_{x,a\sim\beta,r}[\wt{F}_{\text{DI}}(t)] = F(t) + \E_{\Prob}[\Delta]$$
and the variance is 
$$\Var[\wt{F}_{\text{DI}}(t)] =  \frac{1}{n}\Var_{\Prob}\Big[G(t;X,\pi)  + \Delta]. $$ 
\end{lemma}
While the DI estimator has lower variance than both the IS and DR estimators, it suffers from potentially high bias from $\overline{G}$. Unlike the other two estimators, it is biased even when $\wh\beta$ is a perfect estimate of $\beta$, which in practice is undesirable. Though the DI estimator has low bias when $\overline{G}$ is a good model of the condition reward distribution, it is often much easier to form accurate models of $\beta$ than of $G$. 

\subsubsection*{Proofs: Bias and Variance}
We begin by proving the bias and variance expressions of the DR estimator with $\wh\beta$. The bias and variance of the other estimators can be derived as special cases, which we show later.

\begin{proof}[Proof of Lemma~\ref{lem:dr_estim_policy_bias}]
First, we take the expectation of the DR estimator~\eqref{eq:dr_policy_est} with respect to $\Prob_\beta$: 
\begin{align*}
  \E_{\Prob_\beta}\Big[\wt F(t)\Big] &= \E_{\Prob_\beta}\Big[\frac{\pi(A|X)}{\wh\beta(A|X)}\mathbbm{1}\{R \leq t \}\Big] + \E_{\Prob_\beta}\Big[\Big(\frac{\pi(A|X)}{\wh\beta(A|X)} - \sum_a \pi(A|X)\Big) \overline{G}(t;X,A)\Big] \\
  &= \E_{\Prob}\Big[\frac{\beta(A|X)}{\wh\beta(A|X)}\mathbbm{1}\{R \leq t \}\Big] + \E_{\Prob}\Big[\Big(\frac{\beta(A|X)}{\wh\beta(A|X)} - 1\Big) \overline{G}(t;X,A)\Big] \\
  &= F(t) + \E_{\Prob}\Big[\Big(\frac{\beta(A|X)}{\wh\beta(A|X)} - 1\Big)\mathbbm{1}\{R \leq t \}\Big] + \E_{\Prob}\Big[\Big(\frac{\beta(A|X)}{\wh\beta(A|X)} - 1\Big) \overline{G}(t;X,A)\Big] \\
  &= F(t) + \E_{\Prob}\Big[\Big(\frac{\beta(A|X)}{\wh\beta(A|X)} - 1\Big)\Big(G(t;X,A) - \overline{G}(t;X,A)\Big)\Big] \\
  &= F(t) + \E_{\Prob}\Big[\delta(A,X)\Delta(t;X,A)\Big]
\end{align*}
When $\wh\beta = \beta$ for all $a,x$, we have $\delta = 0$, giving the unbiasedness of the estimator. 

Starting from the second line of the proof of variance for the DR estimator (Appendix~\ref{proof:dr_bias_variance}), we have
\begin{align*}
    \Var_{\Prob_\beta}\left[ \wt{F}_{\text{DR}}(t) \right] &= \frac{1}{n}\Var_{\Prob_\beta}\left[ \wh{w}(A,X)\left( \mathbbm{1}_{\lbrace R \leq t \rbrace} - \overline{G}(t;X,A)\right) + \overline{G}(t;X,\pi)  \right] \\ 
    &= \frac{1}{n}\E_{\Prob_\beta}\left[ \wh{w}(A,X)^2\sigma^2(t;X,A) \right] \\
    &\quad\quad+ \frac{1}{n}\Var_{ \Prob_\beta}\left[\wh{w}(A,X)\left(G(t;X,A) - \overline{G}(t;X,A)\right) + \overline{G}(t;X,\pi)  \right] \\ 
    &= \frac{1}{n}\E_{\Prob}\left[ \left( \frac{\beta(A,X)}{\wh\beta(A,X)} \right)^2\sigma^2(t;X,A) \right] \\
    &\quad\quad+ \frac{1}{n}\Var_{\Prob_\beta}\left[\E_{\Prob_\beta}\left[ \wh{w}(A,X)\left(G(t;X,A) - \overline{G}(t;X,A)\right) + \overline{G}(t;X,\pi) \vert X\right] \right] \\
    &\quad\quad+ \frac{1}{n}\E_{\Prob_\beta}\left[\Var_{\Prob_\beta}\left[ \wh{w}(A,X)\left(G(t;X,A) - \overline{G}(t;X,A)\right) \vert X \right]\right] \\ 
    &= \frac{1}{n}\E_{\Prob}\left[ \left( 1-\delta(A,X) \right)^2\sigma^2(t;X,A) \right] \\
    &\quad\quad+ \frac{1}{n}\Var_{\Prob}\left[\E_{\Prob}\left[ \delta(A,X)\Delta(t;X,A) + G(t;X,A) \vert X\right] \right] \\
    &\quad\quad+ \frac{1}{n}\E_{\Prob}\left[\Var_{\Prob_\beta}\left[ \wh{w}(A,X)\Delta(t;X,A) \vert X \right]\right] \\
\end{align*}
the second line uses a change of measure in the first term, and the law of total variance conditioned on the context $X$. The third line follows again from change of measure and substituting in the definition of $\delta$ and $\Delta$. 

\end{proof}

Lemma \ref{lem:is_estim_policy_bias} is derived from Lemma \ref{lem:dr_estim_policy_bias} using the fact that the IS estimator is a special case of the DR estimator with $\overline{G} = 0$. 

Lemma \ref{lem:di_bias} is derived from Lemma \ref{lem:dr_estim_policy_bias} by using $\wh\beta \rightarrow \infty$ which means $\wh\weight = 0$, e.g. importance weighting is not used, and $\delta = 1$.

\subsection{CDF and Risk Estimate Error Bounds} 

Theorem~\ref{THM:IW-CDF-ESTIMATED} generalizes the CDF error bounds established for the IS and DR estimators with known behavior policy to the case where $\wh\beta$ is estimated,  
given an additional high-probability guarantee on the quality of $\wh\beta$. 

\begin{theorem}\label{THM:IW-CDF-ESTIMATED}
For the \text{IS} or \text{DR} \CDF estimator $\wt{F}$ that uses estimated weights $\wh{w}(a,x) = \pi(a|x)/\wh\beta(a,x)$, 
given an estimate $\wh\beta$ that is $\epsilon_\beta$-close to the true behavior policy $\beta$, that is 
\[ 
    \sup_{a,x}\vert \beta(a|x) - \wh\beta(a|x) \vert \leq \epsilon_\beta,
\]
we have with probability at least $1-\delta$ that 
\begin{align*}
    \Prob_\beta\left(\sup_{t\in\Real} \left|\wt{F}(t) -  F(t)\right|\leq \epsilon + c\epsilon_\beta \right)\geq  1 - \delta
\end{align*}
where $\epsilon$ is either $\epsilon_{\text{IS}}$ or $\epsilon = \epsilon_{\text{DR}}$ depending the choice of $\wh{F}$, and $c = w_{max} \left(\inf_{a,x}\wh\beta(a|x)\right)^{-1}$. 
\end{theorem}

Similarly, for $L$-Lipschitz risk functionals, 
the general error bound given in Theorem~\ref{thm:confidence_general} 
can be extended to the case of $\wh\beta$ by adding the additional error term from the policy estimation. 
\begin{corollary}\label{estim_policy_error}
For the \text{IS} or \text{DR} \CDF estimator $\wt{F}$ that uses estimated weights $\wh{w}(a,x) = \pi(a|x)/\wh\beta(a,x)$, 
given an estimate $\wh\beta$ that is $\epsilon_\beta$-close to the true behavior policy $\beta$, 
we have with probability at least $1-\delta$ that 
$$
\left\vert \rho(\wt{F}) - \rho(F) \right\vert \leq L\left(\epsilon + c\epsilon_\beta\right)
$$
where $c = w_{max} \left(\inf_{a,x}\wh\beta(a|x)\right)^{-1}$.
\end{corollary}

Note that the error contributed by policy estimation, $c\epsilon_\beta$, is primarily dependent upon two factors. 
First, the quality of $\wh\beta$ estimation determines the magnitude of $\epsilon_\beta$; a poor estimate naturally leads to a higher value of this constant. 
Second, $c$ is a problem-dependent constant proportional to the maximum importance weight $w_{max}$ and the minimum probability of the estimated behavior policy $\inf_{a,x} \wh \beta(a|x)$. 
If $\inf_{a,x} \wh\beta(a|x)$ is particularly small, the error bound is also large. 
This reflects the fact that \CDF estimation can be difficult when the behavior policy places low probability in some area of the context and action space. 

\begin{remark}
When actions and contexts are discrete, and $\wh\beta$ is estimated using empirical averages, standard concentrations for the mean of a random variable can be used to determine $\epsilon_\beta$. 
If $\wh\beta$ is estimated using regression, depending on the estimator $\epsilon_\beta$ can also be determined from concentration inequalities. 
\end{remark} 

v2\subsubsection*{Proofs: Error Bounds}
The proof of these results is given below. 

\begin{proof}[Proof of Theorem~\ref{THM:IW-CDF-ESTIMATED}]
We can decompose the error $\wh{F} - F$ as:   
\begin{align*}
    \sup_t \vert \wt{F}(t) - F(t) \vert &\leq \sup_t \Big( \vert \wh{F}(t) - F(t) \vert + \vert \wt F(t) - \wh F(t)\vert \Big) \\
    &\leq \sup_t \vert \wh{F}(t) - F(t) \vert + \sup_t \vert \wt F(t) - \wh F(t)\vert
\end{align*}
Theorem \ref{thm:is-cdf} gives a bound for the first term, and the bound for the second term bound is given in Lemma \ref{lem:estimated_policy_is} for the IS estimator,
and in Lemma~\ref{lem:estimated_policy_dr} for the DR estimator. 
\end{proof}

\begin{proof}[Proof of Corollary~\ref{estim_policy_error}] 
    This result follows directly from applying the general risk estimation error bound in Theorem~\ref{thm:confidence_general} to the error from Theorem~\ref{THM:IW-CDF-ESTIMATED}. 
\end{proof}

The intermediary lemmas are defined and proved below: 
\begin{lemma}\label{lem:estimated_policy_is}
Suppose that $|\wh\beta(a|x) - \beta(a|x)| \leq \epsilon_\beta$ for all $a,x$ with probability at least $1-\delta$. Then with probability at least $1-\delta$, 
\begin{align*}
    \sup_t \vert \wt F_{\text{IS}}(t) - \wh F_{\text{IS}}(t)\vert \leq c\epsilon_\beta
\end{align*}
where $c = w_{max} \left(\inf_{a,x}\wh\beta(a|x)\right)^{-1}$.
\end{lemma}

\begin{proof}
We can bound the LHS of the lemma statement as follows. 
\begin{align*}
    \sup_t \vert \wt F_{\text{IS}}(t) - \wh F_{\text{IS}}(t)\vert &= \sup_t \Big\vert \frac{1}{n}\sum_{i=1}^n (w(a_i, x_i) - \wh w(a_i, x_i)) \mathbbm{1}_{\{r_i \leq t\}} \Big\vert \\
    &= \sup_t \Big\vert \frac{1}{n}\sum_{i=1}^n \Big(\frac{\pi(a_i|x_i)}{\beta(a_i|x_i)} - \frac{\pi(a_i|x_i)}{\wh\beta(a_i|x_i)} \Big) \mathbbm{1}_{\{r_i \leq t\}} \Big\vert \\
    &\leq \frac{1}{n}\sum_{i=1}^n\Big\vert \frac{\pi(a_i|x_i)}{\beta(a_i|x_i)} - \frac{\pi(a_i|x_i)}{\wh\beta(a_i|x_i)} \Big\vert\\
    &\leq w_{max}\frac{1}{n}\sum_{i=1}^n \left\vert 1 - \frac{\beta(a_i|x_i)}{\wh\beta(a_i|x_i)} \right\vert \\ 
    &=  w_{max}\frac{1}{n}\sum_{i=1}^n \left\vert \frac{\wh\beta(a_i|x_i) - \beta(a_i|x_i)}{\wh\beta(a_i|x_i)} \right\vert \\ 
    &\leq w_{max}\left(\inf_{a,x}\wh\beta(a|x)\right)^{-1}\frac{1}{n}\sum_{i=1}^n \left\vert \wh\beta(a_i|x_i) - \beta(a_i|x_i) \right\vert\\
    &\leq w_{max}\left(\inf_{a,x}\wh\beta(a|x)\right)^{-1}\epsilon_\beta
\end{align*}
where the last line follows from using the assumption that $|\wh\beta(a|x) - \beta(a|x)| \leq \epsilon_\beta$ for all $a,x$.
\end{proof}

\begin{lemma}\label{lem:estimated_policy_dr}
Suppose that $|\wh\beta(a|x) - \beta(a|x)| \leq \epsilon_\beta$ for all $a,x$ with probability at least $1-\delta$. Then with probability at least $1-\delta$, 
\begin{align*}
    \sup_t \vert \wt F_{\text{DR}}(t) - \wh F_{\text{DR}}(t)\vert \leq c\epsilon_\beta
\end{align*}
where $c = w_{max} \left(\inf_{a,x}\wh\beta(a|x)\right)^{-1}$.
\end{lemma}

\begin{proof}
We can bound the LHS of the lemma statement as follows. Using the definitions of the DR estimators, 
    \begin{align*}
        \sup_t \vert \wt F_{\text{DR}}(t) - \wh F_{\text{DR}}(t)\vert &= \sup_t \Big\vert \frac{1}{n}\sum_{i=1}^n (w(a_i, x_i) - \wh w(a_i, x_i)) \left( \mathbbm{1}_{\{r_i \leq t\}} - \wb{G}(t;x_i,a_i)\right) \Big\vert \\
        &= \sup_t \Big\vert \frac{1}{n}\sum_{i=1}^n \Big(\frac{\pi(a_i|x_i)}{\beta(a_i|x_i)} - \frac{\pi(a_i|x_i)}{\wh\beta(a_i|x_i)} \Big) \left( \mathbbm{1}_{\{r_i \leq t\}} - \wb{G}(t;x_i,a_i)\right) \Big\vert \\
        &\leq \frac{1}{n}\sum_{i=1}^n\Big\vert \frac{\pi(a_i|x_i)}{\beta(a_i|x_i)} - \frac{\pi(a_i|x_i)}{\wh\beta(a_i|x_i)} \Big\vert\\
        &\leq w_{max}\frac{1}{n}\sum_{i=1}^n \left\vert 1 - \frac{\beta(a_i|x_i)}{\wh\beta(a_i|x_i)} \right\vert \\ 
        &=  w_{max}\frac{1}{n}\sum_{i=1}^n \left\vert \frac{\wh\beta(a_i|x_i) - \beta(a_i|x_i)}{\wh\beta(a_i|x_i)} \right\vert \\ 
        &\leq w_{max}\left(\inf_{a,x}\wh\beta(a|x)\right)^{-1}\frac{1}{n}\sum_{i=1}^n \left\vert \wh\beta(a_i|x_i) - \beta(a_i|x_i) \right\vert\\
        &\leq w_{max}\left(\inf_{a,x}\wh\beta(a|x)\right)^{-1}\epsilon_\beta
\end{align*}
where the last line uses the assumption that $|\wh\beta(a|x) - \beta(a|x)| \leq \epsilon_\beta$ for all $a,x$.
\end{proof}

\newpage 
\section{Additional Experiments}\label{appendix:experiment}

\paragraph{Implementation Details. }
Following \cite{dudik2011doubly, dudik2014doubly, wang2017optimal}, 
we obtain our off-policy contextual bandit datasets 
by transforming classification datasets.
The contexts are the provided features, 
and the actions correspond to the possible class labels. 
To obtain the evaluation policy $\pi$,
we use the output probabilities 
of a trained logistic regression classifier \cite{scikit-learn}. 
The behavior policy is defined as 
$\beta = \alpha \pi + (1-\alpha)\pi_{\text{UNIF}}$, 
where $\pi_{\text{UNIF}}$ 
is a uniform policy over the actions,
for some $\alpha \in (0, 1]$. 
Each dataset is generated by drawing actions for each context according to the probabilities of $\beta$, 
and the deterministic reward is 1 if the action matches the ground truth label, and 0 otherwise. 

We apply this process to the set of 9 UCI datasets~\cite{Dua:2019} used in \cite{dudik2011doubly, dudik2014doubly, wang2017optimal}, 
which each have differing dimensions $d$, actions $k$, and sample size $n$. 
Models $\overline{G}$ must be constructed for the $\DM$ and $\DR$ estimators. As in \cite{dudik2011doubly}, 
the dataset is divided into two splits, 
with each of the two splits 
used to estimate $\overline{G}$, 
which is then used 
with the other split 
to calculate the estimator. 
The two results are averaged 
to produce the final estimators. 
In order to estimate $\overline{G}$, 
we discretize the reward support into $t \in [0, 1]$, and train a logistic regression classifier \cite{scikit-learn} for each action $a$ and each $t$, with regularization parameter $C = 1$ and tolerance $0.0001$. The code to reproduce these experiments is provided in the supplementary. On a CPU, they take roughly half a day of compute in total.

\begin{figure}[H]
    \centering
    \includegraphics[width=0.5\textwidth]{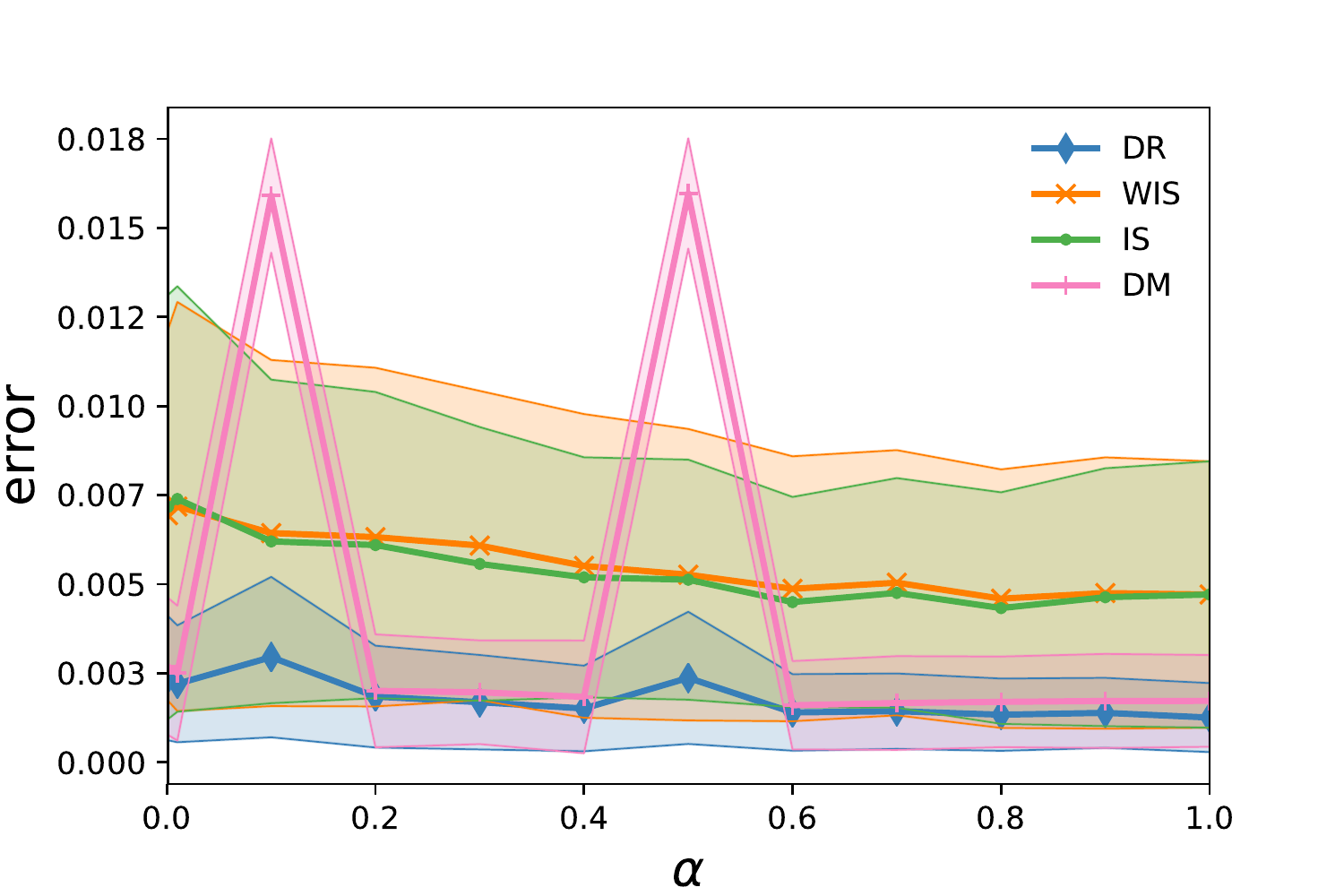}
    \caption{Sup-norm CDF error over $\alpha$ for PageBlocks. Shaded region shows one empirical standard deviation.}
    \label{fig:alpha_cdf_error}
\end{figure}

\begin{figure}[H]
    \centering
    \includegraphics[width=\textwidth]{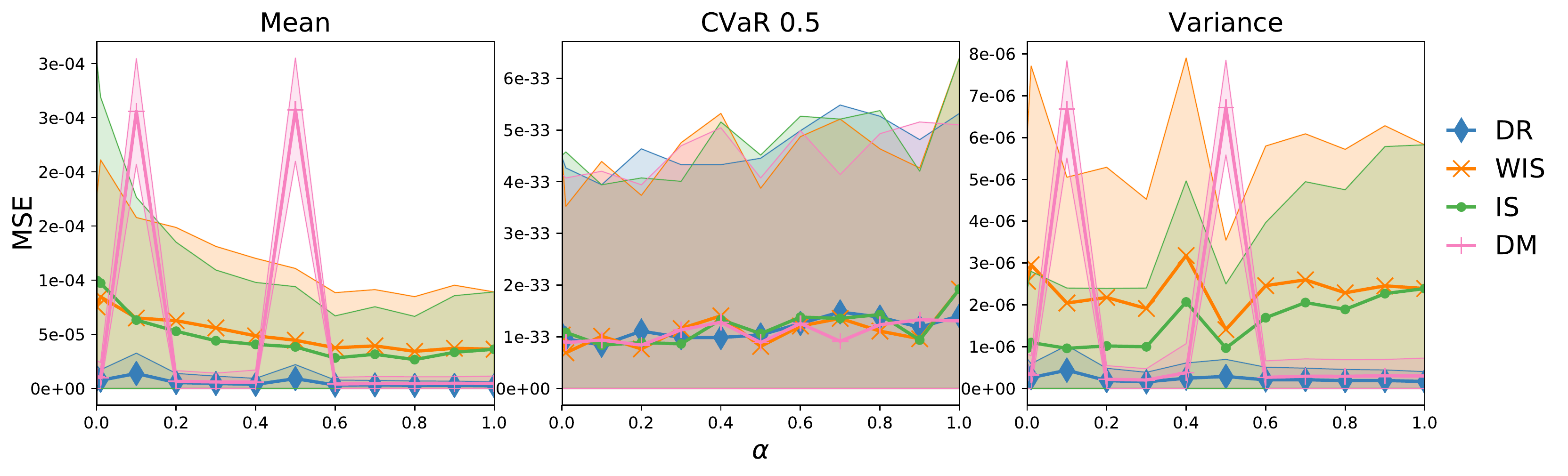}
    \caption{Mean squared error (MSE) over $\alpha$ for different risk functionals evaluated in the PageBlocks dataset. Shaded region shows one empirical standard deviation.  }
    \label{fig:alpha_risk_error}
\end{figure}

\paragraph{Relationship With $\mathbf{\alpha}$. }
We plot the error over the range of $\alpha$, which controls the mismatch between the behavioral policy $\beta$ and the target policy $\pi$ and is thus proportional to $w_{max}$, for the PageBlocks dataset (also in Figure~\ref{fig:cdf_error}). The \CDF error is shown in Figure~\ref{fig:alpha_cdf_error} and the mean squared error (MSE) for the mean, CVaR 0.5, and variance risk functionals are shown in Figure~\ref{fig:alpha_risk_error}. 

The \DR estimator exhibits lower error than any other estimator, and significantly lower variance than the \IS and \WIS estimators, across the range of $\alpha$. This is particularly obvious in the region where $\alpha$ is small, which is where importance weights can become larger and the \IS-based estimators are prone to higher variance. Note that the $\text{CVaR}_{0.5}$ MSE is close to 0 for all estimators. 

\paragraph{Evaluation Over UCI Datasets. } We display the sup-norm error of the estimated \CDF and the mean-squared error (MSE) of estimated risk functionals (mean, $\text{CVaR}_{0.5}$, and variance) for the 9 UCI datasets below. Here, $\alpha = 0.5$ is fixed. All plots are shown over 500 repetitions, with error bars omitted for readability but similar to those shown in Figure~\ref{fig:cdf_error}. 

The general trends reflect analysis presented in Section~\ref{sec:experiments}. 
As expected of our distribution-based approach, trends in \CDF estimation performance are reflected in risk estimation performance. 
Both the \DR and \IS estimators exhibit the expected $O(1/\sqrt{n})$ error convergence across the estimation tasks.
Generally, the \DR estimator does as well as if not better than the other estimators; where the model is difficult to specify well, the \DR estimator may suffer slightly in performance in the low sample regime, but always outperforms the other estimators as the number of samples $n$ increases. 

\begin{figure}[H]
    \centering
    \includegraphics[width=\linewidth]{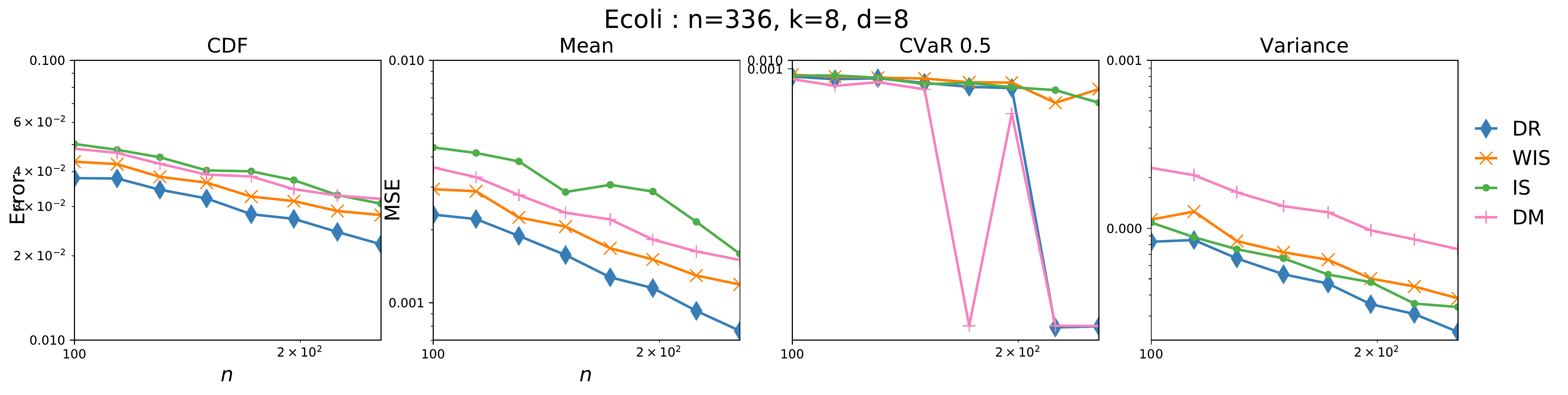} \\
    \includegraphics[width=\linewidth]{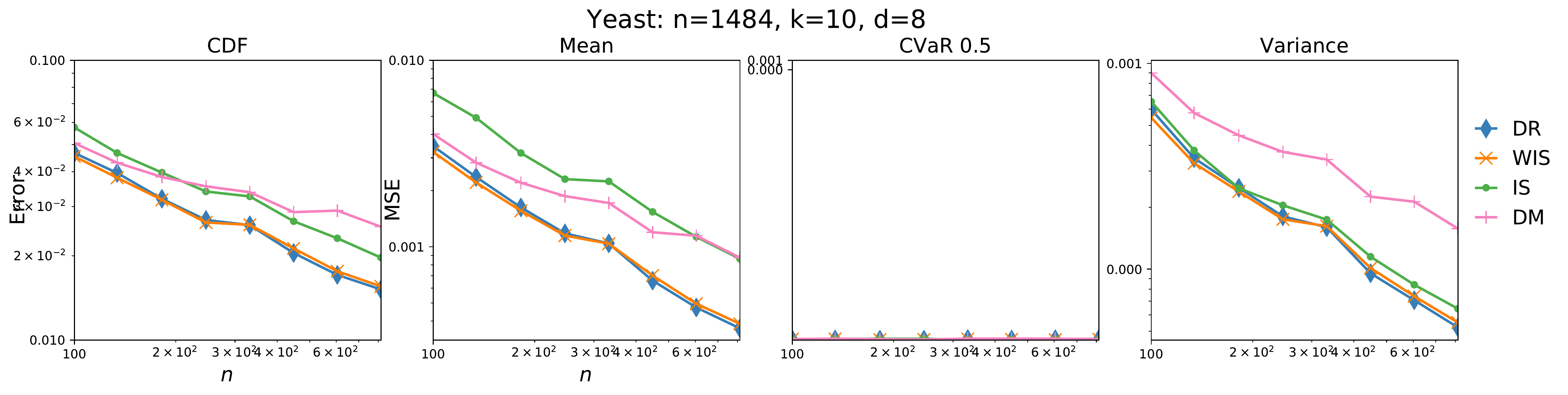} \\
    \includegraphics[width=\linewidth]{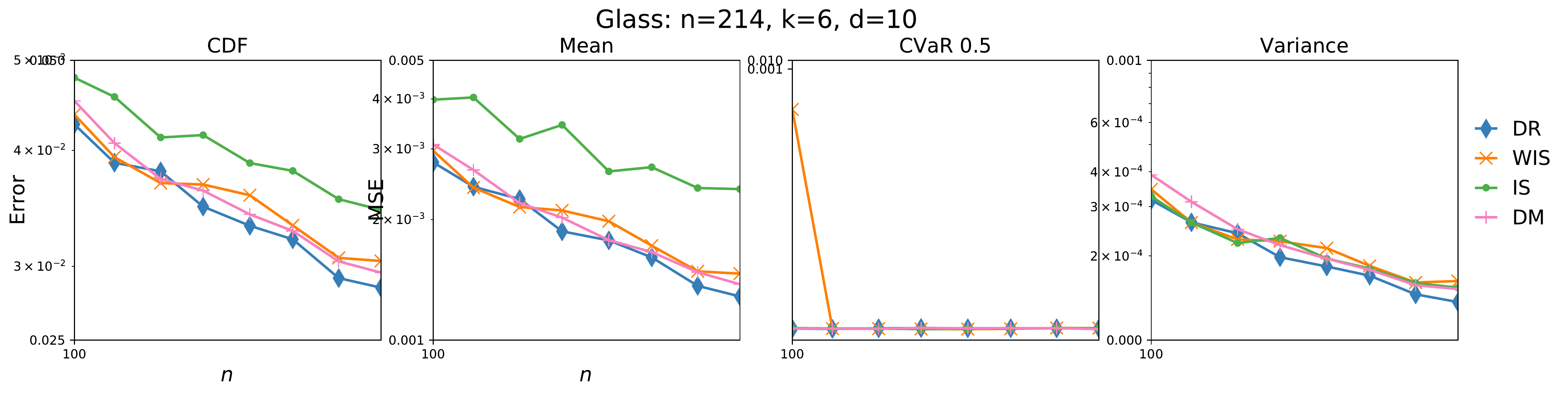}\\
    \includegraphics[width=\linewidth]{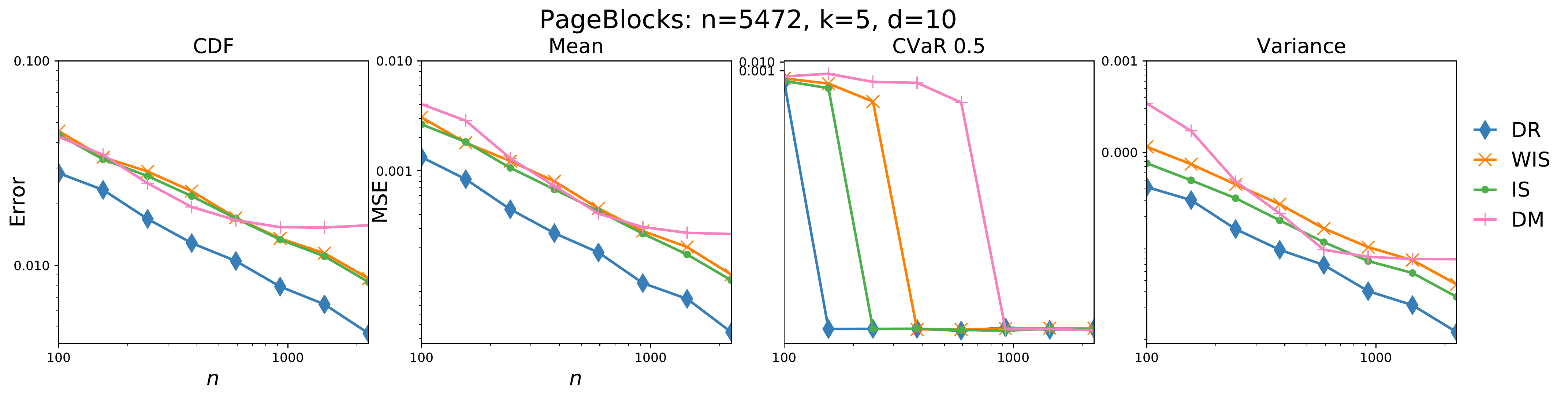}\\
    \includegraphics[width=\linewidth]{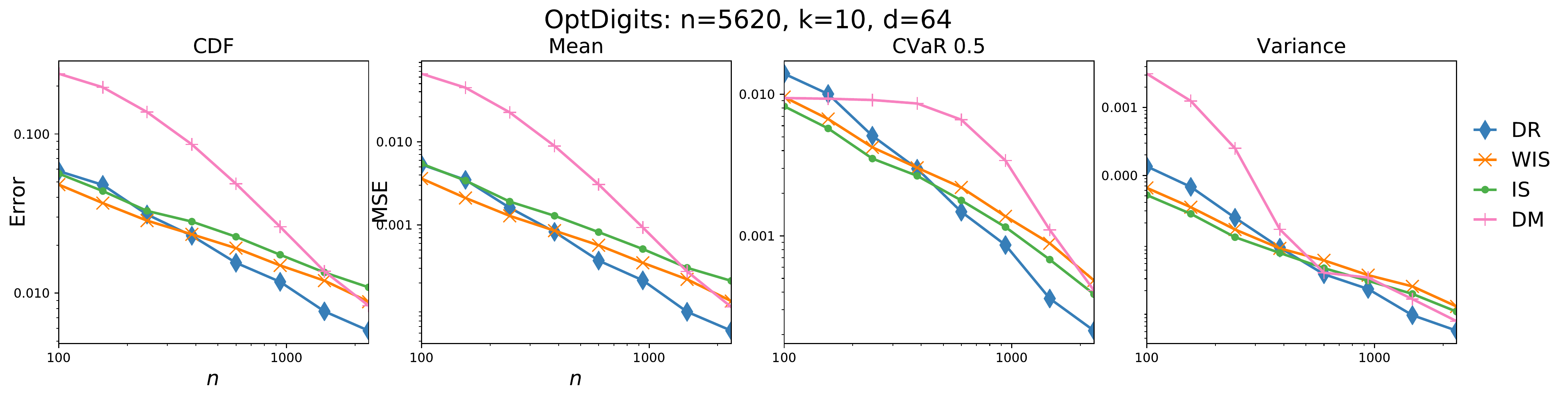} \\
\end{figure}

\begin{figure}[H]
    \centering
     \includegraphics[width=\linewidth]{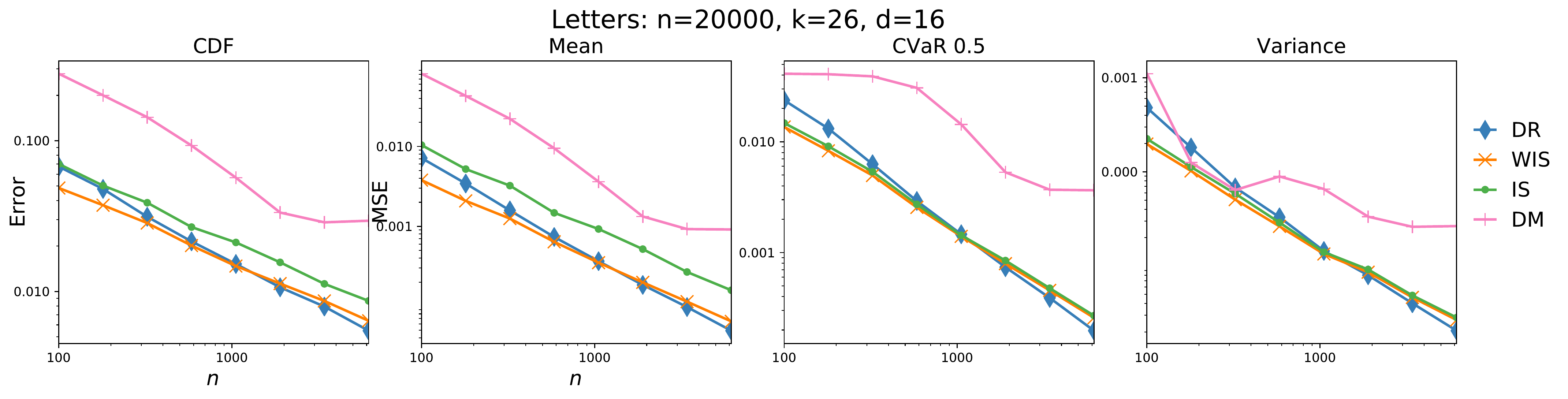} \\
    \includegraphics[width=\linewidth]{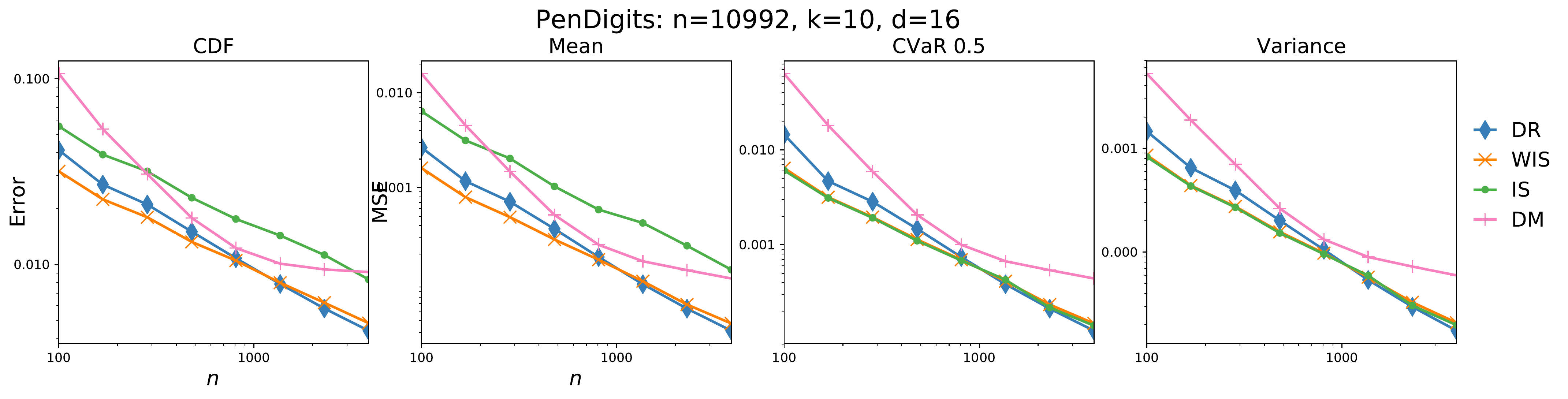} \\
    \includegraphics[width=\linewidth]{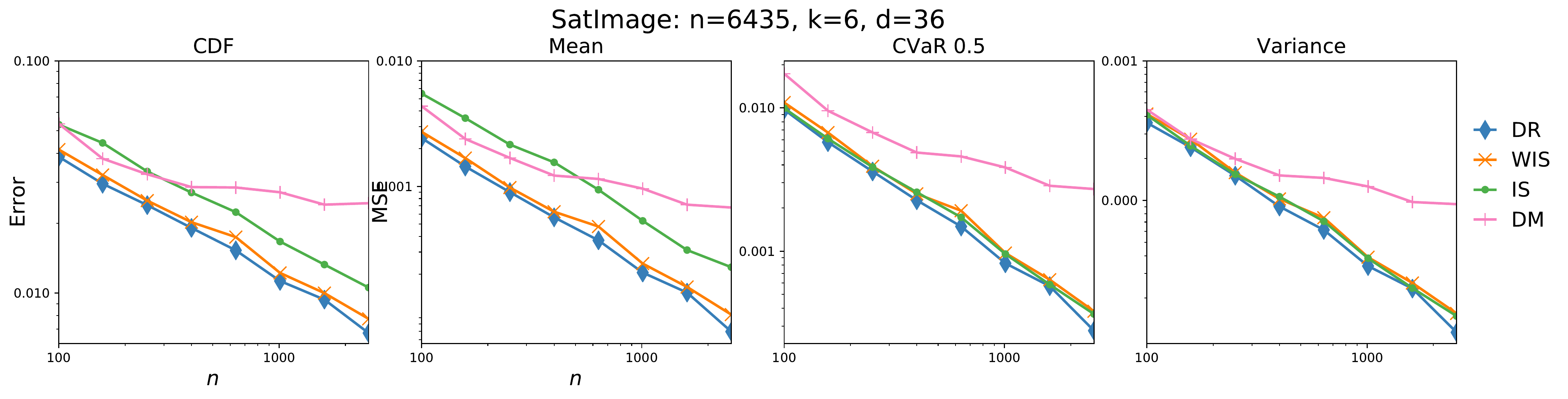} \\
    \includegraphics[width=\linewidth]{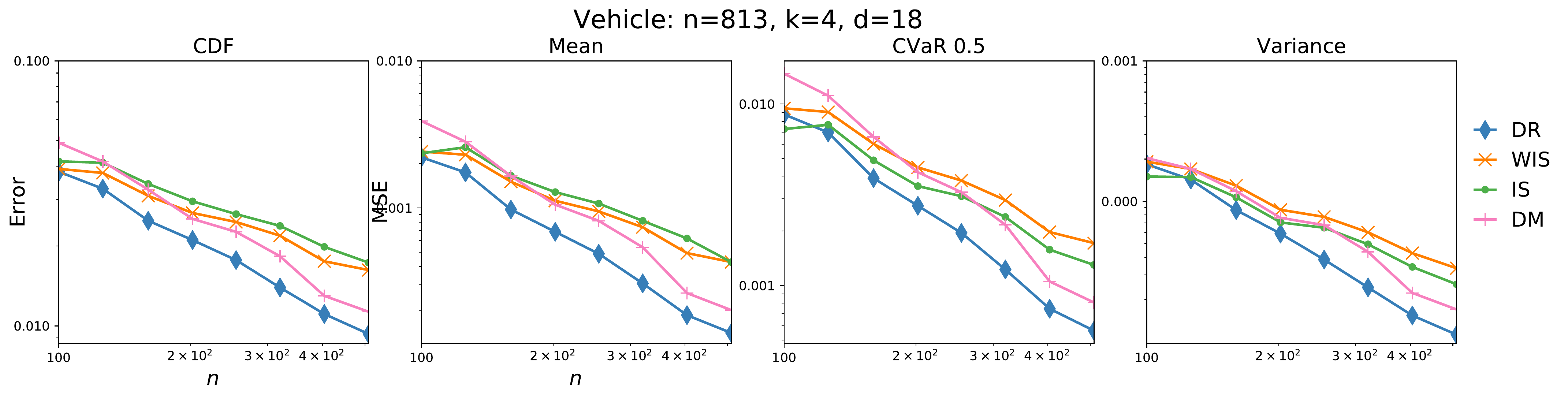}
\end{figure}

\end{document}